\documentclass{article}

\usepackage{iclr2022_conference}

\usepackage{times}
\usepackage[utf8]{inputenc} 
\usepackage[T1]{fontenc}    
\usepackage{hyperref}       
\usepackage{url}            
\usepackage{booktabs}       
\usepackage{amsfonts}       
\usepackage{nicefrac}       
\usepackage{microtype}      
\usepackage{xcolor}         
\usepackage{amsmath}
\usepackage{amssymb}
\usepackage{commath}

\usepackage{amsmath,amsfonts,bm,bbm}

\def\eqref#1{equation~\ref{#1}}

\def\1{\bm{1}}

\DeclareMathAlphabet{\mathsfit}{\encodingdefault}{\sfdefault}{m}{sl}
\SetMathAlphabet{\mathsfit}{bold}{\encodingdefault}{\sfdefault}{bx}{n}

\DeclareMathOperator*{\argmax}{arg\,max}

\newtheorem{theorem}{Theorem}[section]
\newtheorem{corollary}{Corollary}[theorem]

\newtheorem{proof}{Proof}
\newtheorem{definition}{Definition}
\newtheorem{proposition}{Proposition}

\renewcommand{\vec}[1]{\ensuremath{\bm{#1}}}

\newcommand{\mat}[1]{\ensuremath{\mathbf{#1}}}

\usepackage{listings}
\usepackage[frozencache,cachedir=minted]{minted}
\usepackage{color}
\usepackage{todonotes}
\usepackage{wrapfig}
\usepackage{multirow}
\usepackage{enumitem}
\usepackage{cleveref}
\usepackage{subfig}
\usepackage{mathrsfs}
\usepackage{multirow}
\usepackage[ruled,vlined, linesnumbered, noend]{algorithm2e}

\hypersetup{
    colorlinks=true,
    linkcolor = blue,
    anchorcolor = blue,
    citecolor = blue,
    filecolor = blue,
    urlcolor = blue
}

\iclrfinalcopy

\title{\centering Cross-Trajectory Representation Learning for Zero-Shot Generalization in RL}

\author{
  Bogdan Mazoure\thanks{Equal contribution. $^\dagger$The author did part of the work for this paper while at Microsoft.}  $\;^\dagger$ \\
  \texttt{bogdan.mazoure@mail.mcgill.ca} \\
  McGill University, Quebec AI Institute 
  \And
   Ahmed M. Ahmed$^{*\dagger}$\hspace{0.40in} \\
   \texttt{ahmedah@stanford.edu}\hspace{0.40in} \\
   Stanford University\hspace{0.40in} 
   \AND 
   \hspace{0.135in}Patrick MacAlpine$^\dagger$\\
   \hspace{0.135in}\texttt{patrick.macalpine@sony.com} \\
   \hspace{0.135in}Sony AI
   \And 
   R Devon Hjelm\\
   \texttt{devon.hjelm@microsoft.com}\\
   Université de Montréal, Quebec AI Institute, \\
   Microsoft Research
   \And 
   \hspace{1.75in}Andrey Kolobov\\
   \hspace{1.75in}\texttt{akolobov@microsoft.com}\\
   \hspace{1.75in}Microsoft Research
}

\newcommand\longmethod{Cross Trajectory Representation Learning}
\newcommand\shortmethod{CTRL}

\newcommand{\comment}[1]{}

\Crefname{algocf}{Algorithm}{Algorithms}

\begin{document}

\maketitle

\begin{abstract}
A highly desirable property of a reinforcement learning (RL) agent -- and a major difficulty for deep RL approaches -- is the ability to generalize policies learned on a few tasks over a high-dimensional observation space to similar tasks not seen during training. Many promising approaches to this challenge consider RL as a process of training two functions simultaneously: a complex nonlinear \emph{encoder} that maps high-dimensional observations to a latent \emph{representation} space, and a simple linear \emph{policy} over this space. We posit that a superior encoder for zero-shot generalization in RL can be trained by using solely an auxiliary SSL objective if the training process encourages the encoder to map \emph{behaviorally similar} observations to similar representations, as reward-based signal can cause overfitting in the encoder \citep{raileanu2021decoupling}. We propose \longmethod\ (\shortmethod), a method that runs within an RL agent and conditions its encoder to recognize behavioral similarity in observations by applying a novel SSL objective to pairs of trajectories from the agent's policies. \shortmethod\ can be viewed as having the same effect as inducing a pseudo-bisimulation metric but, crucially, avoids the use of rewards and associated overfitting risks. Our experiments\footnote{Code link: \url{https://github.com/bmazoure/ctrl_public}} ablate various components of \shortmethod\ and demonstrate that in combination with PPO it achieves better generalization performance on the challenging Procgen benchmark suite \citep{cobbe2020leveraging}.
\end{abstract}

\section{Introduction}
Deep reinforcement learning (RL) has emerged as a powerful tool for building decision-making agents for domains with high-dimensional observation spaces, such as video games~\citep{mnih2015humanlevel}, robotic manipulation~\citep{levine2016endtoend}, and autonomous driving~\citep{kendall2019driving}. 
However, while deep RL agents may excel at the specific task variations they are trained on, learning behaviors that generalize across a large family of similar tasks, such as handling a variety of objects with a robotic manipulator, driving under a variety of conditions, or coping with different levels in a game, remains a challenge. 
This problem is especially acute in zero-shot generalization (ZSG) settings, where only a few sequential tasks are available to learn policies that are meant to perform well on different yet related tasks without further parameter adaptation.
ZSG settings highlight the fact that generalization often cannot be solved by more training, as it can be too expensive or impossible to instantiate all possible real-world deployment scenarios a-priori.

In this work, we aim to improve ZSG in RL by proposing a new way of training the agent's \emph{representation}, a low-dimensional summary of information relevant to decision-making extracted from the agent's high-dimensional observations. 
Outside of RL, representation learning can help with ZSG, e.g. using unsupervised learning to obtain a representation that readily transfers to unseen classes in vision tasks~\citep{bucher2017generating, sylvain2019locality, wu2020self}.
In RL, unsupervised representation learning in the form of auxiliary objectives can be used to provide a richer learning signal over learning from reward alone, which helps the agent avoid overfitting on task-specific information~\citep{raileanu2021decoupling}.
However, to our knowledge, no unsupervised learning method used in this way in RL has thus far been shown to substantially improve performance in ZSG over end-to-end reward-based methods~\citep[e.g.,][]{cobbe2020leveraging}.

We posit that using unsupervised (reward-free) learning to find representations that capture behavioral similarity across different trajectories will improve ZSG in RL.
We note that the bisimulation framework~\citep{ferns2004metrics} does this directly with rewards, optimizing an agent to treat states as behaviorally similar based on the expected reward, and this has been shown to help in visual generalization settings~\citep{zhang2020learning}. 
We expand on this framework to improve ZSG performance, using unsupervised learning to train an agent that recognizes behavior similarity in a reward-free fashion.
To do so, we propose \emph{\longmethod{} (\shortmethod)}, which applies a novel self-supervised learning (SSL) objective to pairs of trajectories drawn from the agent's policies. 
For optimization, \shortmethod\ defines a prediction objective across trajectory representations from nearby partitions defined by an online clustering algorithm.
The end result is an agent whose encoder maps behaviorally similar trajectories to similar representations without directly referencing reward, which we show improves ZSG performance over using pure RL or RL in conjunction with other unsupervised or SSL methods.

Our main contributions are as follows: 
\begin{itemize}
\item We introduce \longmethod~(\shortmethod), a novel SSL algorithm for RL that defines an auxiliary objective across trajectories in order to capture the notion of \emph{behavioral similarity} in the representations of the agent's belief states. \shortmethod's approach is two-fold: (i) it uses a clustering loss to group representations of behaviorally similar trajectories and (ii) boosts cross-predictivity between trajectory representations from nearby clusters. 

\item We empirically show that \shortmethod\ improves zero-shot generalization in the challenging Procgen benchmark suite~\citep{cobbe2020leveraging}. Through a series of ablations, we highlight the importance of cross-trajectory views in boosting behavioral similarity.
\item We connect {\shortmethod} to the class of bisimulation methods, and provide sufficient conditions under which both formalisms can be equivalent.
\end{itemize}

\section{Background, motivation, and related works}\label{sec:rel_work}

There are a broad class of ZSG settings in RL, such as generalization across reward functions~\citep{barreto2016successor,touati2021learning,misra2020kinematic}, observation spaces~\citep{zhang2020learning,li2021domain,raileanu2021decoupling}, or task dynamics~\citep{rakelly2019efficient}.
For each of these settings, there are a number of promising directions for improving ZSG performance: giving the agent better exploration policies~\citep{van2014generalization, misra2020kinematic, agarwal2020pc}, meta learning~\citep{oh2017zero,gupta2018unsupervised,rakelly2019efficient}, or planning~\citep{sohn2018hierarchical}.
In this work, we focus on directly improving the agent's \emph{representations}.
The agent's representations are high-level abstractions of observations or trajectories from the environment (e.g., the output of an encoder), and the desired property here is that one can easily learn a policy on top of that representation such that the combined model (i.e., the agent) generalizes to novel situations.
The tasks are assumed to share a common high-level goal and are set in environments that have the same dynamics, but each task may need to be accomplished under different initial conditions and may differ visually.
As the policy is built upon the agent's representations, this motivates the focus of this work for improving generalization: unless the agent's representations generalize well, one cannot expect its policy to readily do so.

Unsupervised representation learning has been shown to improve generalization across domains, including zero-shot in vision~\citep{sylvain2019locality, wu2020self} and sample-efficiency in RL~\citep{eysenbach2018diversity, schwarzer2020dataefficient, stooke2020decoupling}. In RL, unsupervised objectives can be used as an \emph{auxiliary objective}~\citep[or auxiliary task,][]{jaderberg2016reinforcement}, which provide an alternative signal to reward-based learning signal.
Due to the potential role the RL loss may play in overfitting~\citep{raileanu2021decoupling}, we believe that having a learning objective for agent's representation that is separate from that of its policy is crucial for good ZSG performance.

\paragraph{Self-supervised learning and reinforcement learning.}
A successful class of models that incorporate unsupervised objectives to improve RL use self-supervised learning (SSL)~\citep{anand2019unsupervised, srinivas2020curl, mazoure2020deep, schwarzer2020dataefficient, stooke2020decoupling, higgins2017darla}.
SSL formulates objectives by generating different \emph{views} of the data, which are essentially transformed versions of the data, e.g., generated by using data augmentation or by sampling patches.
While successful in their own way, prior works that combine SSL with RL do so by applying known SSL algorithms~\citep[e.g., from vision,][]{hjelm2018learning, oord2019representation, bachman2019learning, chen2020simple, he2020momentum, grill2020bootstrap} to RL in a nearly off-the-shelf manner, predicting state representations \textit{within} a given trajectory, only potentially using other trajectories as counterexamples in a contrastive loss. As such, these methods' representations can have trouble generalizing latent behavioral patterns present in ostensibly different trajectories.

\paragraph{Bisimulation metrics in reinforcement learning.}
Our hypothesis is that ZSG is achievable if the agent recognizes behavioral similarity between trajectories based on their long-term evolution.
Learning this sort of behavior similarity is a central characteristic of bisimulation metrics~\citep{ferns2004metrics}, which assign a value of 0 to states which are behaviorally indistinguishable and have the same reward.
Reward-based bisimulation metrics have been shown to learn representations that have a number of useful properties, e.g.: smoothness~\citep{gelada2019deepmdp}, visual invariance~\citep{zhang2020learning}, action equivariance~\citep{vanderpol2020plannable} and multi-task adaptation~\citep{zhang2020learning2}.
For ZSG however, encoding relational information based on reward may not actually help~\citep{misra2020kinematic, touati2021learning, yang2021representation, agarwal2021contrastive}, as the agent may overfit to spurious correlations between high-dimensional observations and the reward signal seen during training.
HOMER~\citep{misra2020kinematic} expands on the concept of bisimulation to learn behavioral similarity between states using unsupervised exploration at deployment. Among the existing methods, PSEs~\citep{agarwal2021contrastive} reward-free notion of behavioral similarity is conceptually the closest to \shortmethod's, and we compare these algorithms empirically in \Cref{sec:results}. However, algorithmically and in terms of their modes of operation, \shortmethod\ and PSE are very different. PSE assumes the availability of expert policies for training tasks and learns a representation using trajectories from these experts and an action distance measure, which it also assumes to be provided. \shortmethod\ doesn't make these assumptions and learns a representation online from trajectories simultaneously generated by its substrate RL algorithm.

\paragraph{Mining views across unsupervised clusters.}
Given our hypothesis that a model that learns behavioral similarities using signal other than reward will perform well on ZSG, there are still many potential models available to learn said similarities in an unsupervised way.
A simple and natural choice is to collect agent trajectories as examples of behaviors, then do clustering \citep[online, similar to][]{asano2020self, caron2020unsupervised} over trajectories. 
In RL, Proto-RL~\citep{yarats2021reinforcement} also uses clustering to obtain a pre-trained set of prototypical states, but for a different purpose -- to estimate state visitation entropy in hard exploration problems.
However, clustering alone may not be sufficient to recognize behaviors necessary for ZSG, as representations built on clustering only need to partition behaviors, which may bias the model towards similarities evident training experience. This would be counter-productive for our generalization goal. 
We therefore use a second objective built on top of the structure provided by clustering to learn a more diverse set of similarities. 
Drawing inspiration from Mine Your Own View~\citep[MYOW,][]{azabou2021view}, \shortmethod{} selects (\emph{mines}) representational nearest neighbors from different, nearby clusters and applies a predictive SSL objective to them. 
This cross-cluster objective encourages \shortmethod{} to recognize a larger set of similarities than would be necessary to cluster on the training set, which we show improves ZSG performance.

\section{Problem statement and preliminaries}\label{sec:problem}

Formally, we define our problem setting w.r.t. a discrete-time Markov decision process (MDP) $M \triangleq \langle \mathcal{S}, \mathcal{A}, \mathcal{P}, \mathcal{R} \rangle$, where $\mathcal{S}$ is a state space, $\mathcal{A}$ is an action space, $\mathcal{P}: \mathcal{S} \times \mathcal{A} \times \mathcal{S} \to [0, 1]$ is a transition function characterizing environment dynamics, and $\mathcal{R}: \mathcal{S} \times \mathcal{A} \to \mathbb{R}$ is a reward function. $M$'s state and action spaces may be discrete or continuous, but in the rest of the paper we assume them to be discrete to simplify exposition. In practice, an agent usually receives observations but not the full information about the environment's current state. Consider an observation space $\mathcal{O}$ and an observation function $\mathcal{Z}: \mathcal{S} \times \mathcal{O} \to [0, 1]$ that define what observations an agent may receive and how these observations are generated (possibly stochastically) from $M$'s states. We define a \emph{task} $T$ as a partially observable MDP (POMDP) $T =  \langle \mathcal{S}, \mathcal{A}, \mathcal{P}, \mathcal{R}, \mathcal{O}, \mathcal{Z}, s_0\rangle$, where $s_0 \in \mathcal{S}$ is an initial state. Although many RL agents make decisions in a POMDP based only on the current observation $o_t$ or at most a few recent ones, in general this may require using information from the entire observation history $o_1, \ldots, o_t$ so far. Denoting the space of such  histories as $\mathcal{H}$, computing an agent's behavior for task $T$ amounts to finding a policy $\pi: \mathcal{H} \times \mathcal{A} \to [0, 1]$ with the optimal or near-optimal expected return from the initial state $V^\pi_T \triangleq \mathbb{E}\left[\sum_{t=0}^\infty \gamma^t \mathcal{R}(S_t, \pi(H_t) \mid S_0 = s_0\right]$, where $S_t$  and $H_t$ are random variables for the POMDP's underlying state and agent's observation history at time step $t$, respectively, and $\gamma$ is a discount factor. For an MDP $M = \langle \mathcal{S}, \mathcal{A}, \mathcal{P}, \mathcal{R} \rangle$, a set $\mathscr{O}$ of observation spaces, and a set $\mathscr{Z}$ of observation functions w.r.t. $\mathcal{S}$, let a \emph{task family} be the POMDPs set $\mathscr{T}_{M, \mathscr{O}, \mathscr{Z}} \triangleq\{\langle \mathcal{S}, \mathcal{A}, \mathcal{P}, \mathcal{R}, \mathcal{O}, \mathcal{Z}, s_0 \rangle \}_{ \mathcal{O} \in \mathscr{O}, \mathcal{Z} \in \mathscr{Z}, s_0 \in \mathcal{S}}$. We assume that different observation spaces in $\mathscr{O}$ have the same mathematical form, e.g., pixel tensors representing possible camera images, but  correspond to qualitatively distinct subspaces of this larger space, such as subspaces of images depicting brightly and dimly lit scenes.

Our training and evaluation protocol is formalized w.r.t. a \emph{task distribution} $d(\mathscr{T}_{M, \mathscr{O}, \mathscr{Z}}) \triangleq P(\mathscr{O},\mathcal{Z}, \mathcal{S})$ \emph{over task family $\mathscr{T}_{M, \mathscr{O}, \mathscr{Z}}$}, where $P(\mathscr{O},\mathcal{Z}, \mathcal{S})$ is a joint probability mass over observation spaces, observation functions, and initial states. In the rest of the paper, $M$, $\mathscr{O}$, and $\mathcal{Z}$ will be clear from context, and we will denote the task family as $\mathscr{T}$ and the task distribution as $d(\mathscr{T})$. For agent training, we choose $N$ tasks from $\mathscr{T}$, and denote the distribution $d(\mathscr{T})$ restricted to these $N$ tasks as $d(\mathscr{T}_N)$. During the training phase, the RL agent learns via a series of epochs (themselves composed of episodes), by sampling a task from $d(\mathscr{T})$ independently at the start of each episode, until the total number of time steps exceeds its training budget. In each epoch, an RL algorithm uses a batch of trajectories gathered from episodes in order to compute gradients of an RL objective and update the parameters of the agent's policy $\pi$. In representation learning-aided RL, a policy is viewed as a composition $\pi = \theta \circ \phi$  of an \emph{encoder} $\phi:\mathcal{H} \to \mathcal{E}$ and a \emph{policy head} $\theta: \mathcal{E} \times \mathcal{A} \to [0,1]$, both of which are the outputs of the training phase. 
The training phase is followed by an evaluation phase, during which the policy is applied to tasks sampled from $d(\mathscr{T}\setminus \mathscr{T}_N)$, distribution $d$ restricted to the set of tasks $\mathscr{T}\setminus \mathscr{T}_N$ not seen during training. \emph{Our focus on generalization means that we seek a policy $\pi$ whose encoder $\phi$ allows it to maximize $\mathbb{E}_{T \sim d(\mathscr{T}\setminus \mathscr{T}_N)}[V^\pi_T]$ despite being trained only on distribution $d(\mathscr{T}_N)$.}

\section{Algorithm}\label{sec:algorithm}

\shortmethod's key conceptual insight is that capturing reward-agnostic behavioral similarity improves ZSG, because it enables $\phi$ to correctly associate previously unseen observation histories with those for which the agent's RL-trained behavior prescribes a good action.
\shortmethod\ runs synchronously with an online RL algorithm, which is crucial to ensure that as the agent's policies improve, so does the notion of behavioral similarity induced by \shortmethod. 

Like most online RL methods themselves, our algorithm operates in epochs, learning from a batch of trajectories in each epoch. CTRL assumes that all trajectories within each of its training batches come from the same policy. Before the RL algorithm updates the policy head in a given epoch, CTRL uses a trajectory batch from the current policy to update the encoder with gradients of a novel auxiliary loss $\mathcal{L}_{\text{\shortmethod}}$ that we describe in this section.

\subsection{Intuition and high-level description}

\noindent
\textbf{Algorithm overview.} For each trajectory batch, CTRL performs 4 operations:

\begin{enumerate}

\item{} Apply the observation history (belief state) encoder $\phi$  to generate a low-dimensional reward-agnostic representation (“view”) of each trajectory.

\item{} Group trajectories’ views into $C$  sets ($C$ is a tunable hyperparameter) using an online clustering algorithm with loss $\mathcal{L}_\text{clust}$ (\Cref{eq:sinkhorn_loss}, \Cref{sec:details}).

\item{} Using trajectory pairs selected from neighboring clusters, apply a predictive loss $\mathcal{L}_\text{pred}$ (\Cref{eq:pred_loss}, \Cref{sec:details}) to encourage $\phi$ to capture cross-cluster behavioral similarities.

\item{} Update $\phi$ with gradients of the total loss: $\mathcal{L}_{\text{\shortmethod}} = \mathcal{L}_{\text{clust}} + \mathcal{L}_{\text{pred}}$.

\end{enumerate}

\begin{figure}
    \centering
    \includegraphics[width=\linewidth]{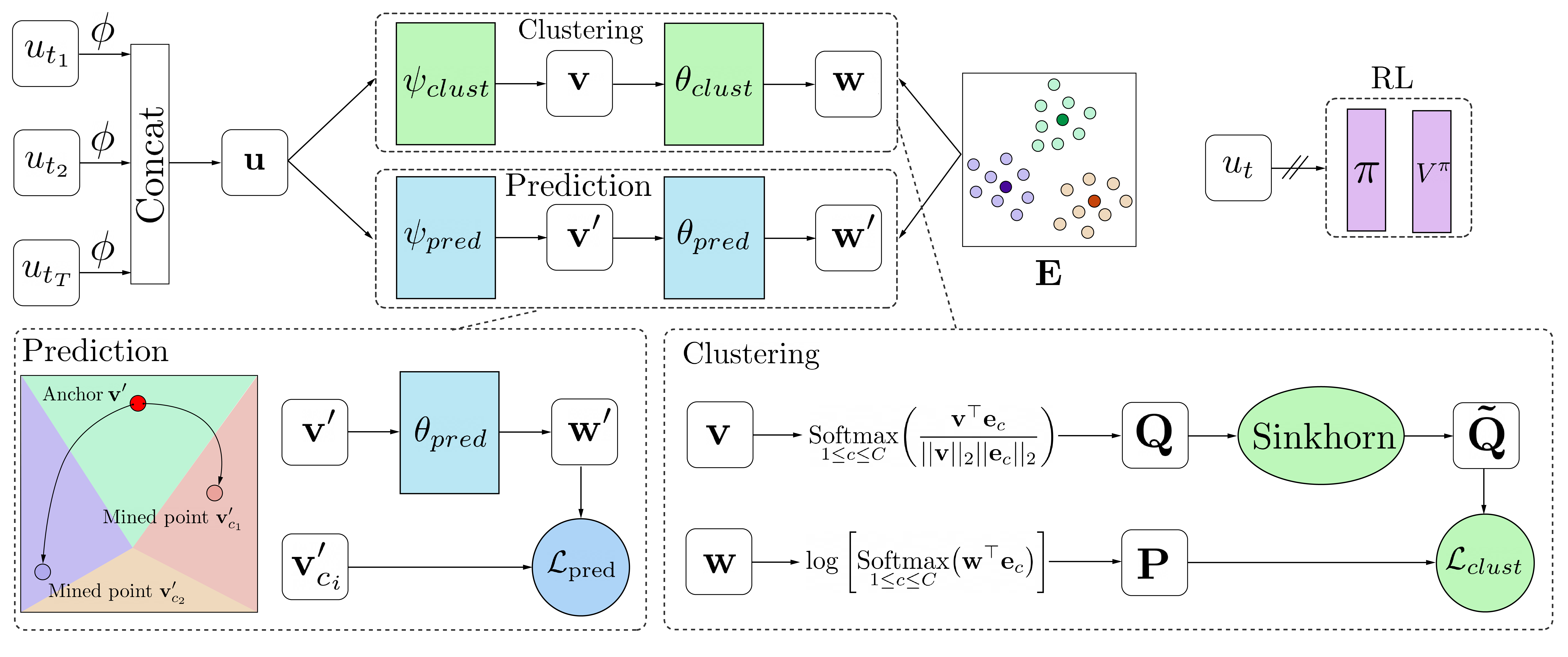}
    \caption{\small Schematic view of \shortmethod's key steps for every trajectory batch. \textbf{(i) Generating trajectory views \textit{(top left)}}. For each trajectory in a batch, \shortmethod\ samples a subsequence of its time steps, computes belief-state/action embeddings $u_{t_i}$ with encoder $\phi$, and concatenates them into a trajectory representation (view) $\mathbf{u}$. \textbf{(ii) Clustering trajectory views \textit{(bottom right)}.}  \shortmethod\ uses the online Sinkhorn-Knopp clustering procedure~\citep{caron2020unsupervised}: for each trajectory view $u$, it produces two new views $\mathbf{v}$ and $\mathbf{w}$, soft-clusters all trajectories' $\mathbf{v}$s and $\mathbf{w}$s into $C$ clusters, and uses a measure of consistency between these two clusterings as a loss $\mathcal{L}_{\text{clust}}$. In the diagram, variables $\mathbf{e}_c$ denote cluster centroids. \textbf{(iii) Encouraging cross-cluster behavioral similarity \textit{(bottom left)}.} After computing trajectory view clusters, \shortmethod\ applies a variant of MYOW~\citep{azabou2021view} to them. Namely, it repeatedly samples a trajectory view $\mathbf{v}'$, computes a new view $\mathbf{w}'$ for it, and computes a loss $\mathcal{L}_{\text{pred}}$ that penalizes differences between $\mathbf{w}'$ and views $\mathbf{v'_{c_i}}$ of randomly chosen trajectories from $\mathbf{v}'$'s neighboring clusters. Encoder $\phi$ and auxiliary predictors used by \shortmethod\ are then updated using $\mathcal{L}_{\text{\shortmethod}} = \mathcal{L}_{\text{clust}} + \mathcal{L}_{\text{pred}}$'s gradients  \textbf{\textit{(top right)}}.}
    \label{fig:ctrl}
\end{figure}

The schema in \Cref{fig:ctrl} provides a high-level outline of these steps' implementation and explains their interplay within \shortmethod, accompanied by an intuition for each step (below) a more detailed description in \Cref{sec:details}. We conduct ablations to show the effect of removing the clustering and predictive objectives of \shortmethod, with details in Appendix~\ref{sec:ablations_algo}.

\textbf{Clustering.}  $C$ clusters can be viewed as corresponding to $C$  latent “situations” in which an RL agent may find itself. Each situation is essentially a group of belief states. \shortmethod’s implicit hypothesis is that a given policy should behave roughly similarly across all belief states corresponding to the same “situation”, i.e., generalize across similar belief states. Under this hypothesis, an agent’s policy can be expected to produce $C$ sets of roughly similar trajectories. CTRL’s clustering step (\#2 above) is an attempt to recover these trajectory sets. Since each trajectory consists of belief states, the purpose of $\mathcal{L}_{\text{clust}}$ is to force the encoder $\phi$ to compute belief state representations that make trajectories within each cluster look similar in the latent space.

Since we would like to evolve clustering online as new trajectory batches arrive, we employ a common online clustering algorithm, the Sinkhorn-Knopp procedure~\citep{caron2020unsupervised}, which has been used as an auxiliary RL loss~\citep[e.g., Proto-RL,][]{yarats2021reinforcement}. 

\textbf{Cross-cluster prediction.} Note, however, that the clustering loss emphasizes the recognition of behavioral similarities \emph{within} clusters.

This may hurt generalization, as the resulting centroids may not faithfully represent behaviors encountered at test time. Our hypothesis is that encouraging encoder $\phi$ to induce latent-space similarities between trajectories from \emph{different but adjacent} clusters will increase its ability to recognize behaviors in unseen test trajectories.

While there are several ways to encourage cross-cluster representational similarity, using a mechanism similar to MYOW~\citep{azabou2021view} on trajectories drawn from neighboring clusters captures this idea particularly well. Namely, to get the cross-predictive loss $\mathcal{L}_{\text{pred}}$, we sample trajectory view pairs from neighboring clusters and apply the cosine-similarity loss to those pairs.

\textbf{Using reward guidance without reward signal for representation learning.} \shortmethod\ trains encoder $\phi$ only using the gradients of $\mathcal{L}_{\text{\shortmethod}}$; the RL algorithm's loss $\mathcal{L}_{\text{RL}}$ trains only the policy head. Thus, encoder $\phi$ is isolated from the previously observed dangers of overfitting to the reward function~\citep{raileanu2021decoupling} that shapes $\mathcal{L}_{\text{RL}}$. However, we emphasize that \shortmethod's representation learning is nonetheless very much guided by the reward function, although indirectly: the training batches of belief state and action trajectories are still collected from policies learned by the policy head via $\mathcal{L}_{\text{RL}}$'s gradients, which are reward-dependent.

\subsection{Details \label{sec:details}}

Below we describe the details of each of \shortmethod's steps, with \shortmethod's pseudocode presented in \textbf{\Cref{alg:main} in Appendix~\ref{sec:pseudocode}}. While in general the agent's belief state at step $t$ of a trajectory is the entire observation history $h_t = (o_1, \ldots, o_t)$, in the rest of the section we will assume $h_t = (o_t)$ and, in a slight abuse of notation, use $\phi(o_t)$ instead of $\phi(h_t)$ to simplify explanations\footnote{While, in theory, the Procgen suite is indeed a POMDP, most RL algorithms take the most recent observation as the belief state -- a simplification which was shown not to hinder ZSG on Procgen~\citep{cobbe2020leveraging}.}. We emphasize, however, that \shortmethod\ equally applies in settings where the agent uses a much longer history of observations as its state. In this case, $\phi$ would be recurrent or process stacks of frames.

We also note that our \shortmethod\ implementation's high-level algorithmic choices for clustering and cross-cluster prediction -- Sinkhorn-Knopp and MYOW, respectively -- come from prior works~\citep{caron2020unsupervised,azabou2021view}. \\

\textbf{Generating low-dimensional trajectory views with encoder $\phi$.} \shortmethod's input in each epoch is a trajectory batch $\{traj_i\}_{i=1}^B$ of size $B$. Assume all trajectories in the batch have the same length $L$. For an integer hyperparameter $T \leq L$, for each trajectory $traj_i=(o_0, a_0, r_0, \ldots o_{T_b}, a_{L},r_{L})$ we independently and uniformly sample a subset of its steps $\tau_i = t_1, \ldots, t_T$ to form a subtrajectory $(o_{t_1}, a_{t_1}, r_{t_1}, \ldots o_{t_T}, a_{t_T},r_{t_T})$. We then encode this subtrajectory as 
\begin{equation}
\vec{u}_i^{(\tau_i)}=(FiLM(\phi(o_{t_1}),a_{t_1}),..,FiLM(\phi(o_{t_T}),a_{t_T})),
    \label{eq:clustering_tuples}
\end{equation}

where $FiLM(\phi(o_{t_j}),a_{t_j})$ is a common way of combining representations of different objects, akin to conditioning \cite{perez2017film}, and the resulting vector $\vec{u}_i^{(\tau_i)}$ is in a low-dimensional space $\mathcal{U}$. Note two aspects of the process of generating these vectors: (1) it drops rewards from the original trajectory and (2) it critically relies on $\phi$ whose parameter values are learned from previous epochs. Vectors $\vec{u}_i^{(\tau_i)}$ produced in this way are the trajectory views that the next steps of \shortmethod\ operate on.\\

\textbf{Clustering trajectory views.}  \shortmethod\ groups trajectories from the epoch's batch by clustering the set of their views $\{\vec{u}_i^{(\tau_i)}\}_{i=1}^B$.  Since we would like to evolve clustering online as new trajectory batches arrive, we employ a common online clustering algorithm, the Sinkhorn-Knopp procedure~\citep{caron2020unsupervised}, which has been used as an auxiliary RL loss~\citep[e.g., Proto-RL,][]{yarats2021reinforcement}. 
Since \shortmethod\ operates online, Sinkhorn-Knopp is better-suited for the task than other clustering methods.

The clustering branch computes two views of each input $\vec{u}_i^{(\tau_i)}$ in a cascading fashion: first by passing it through a clustering encoder $\psi_{\text{clust}}:\mathcal{U} \to \mathcal{V}$, e.g. an RNN, to obtain a lower-dimensional view  $\vec{v}_i=\psi_{\text{clust}}(\vec{u}_i^{(\tau_i)})$, and then by passing $\vec{v}_i$ through yet another network, an MLP $\theta_{\text{clust}}:\mathcal{V}\to \mathcal{W}$, to produce view $\vec{w}_i=\theta_{\text{clust}}(\vec{v}_i)$. The parameters of $\psi_{\text{clust}}$ and $\theta_{\text{clust}}$ are learned through the epochs jointly with $\phi$'s. Like $\vec{u}_i^{(\tau_i)}$, each $\vec{v}_i$ and $\vec{w}_i$ is a view of trajectory $i$; the approach then consists in projecting $\vec{v}_i$'s and $\vec{w}_i$'s onto centroids of $C$ clusters in two different ways and then computes a clustering loss that enforces consistency between $\vec{v}_i$'s and $\vec{w}_i$'s cluster projections.

Specifically, we represent the centroid of each cluster $c$ with a vector $\vec{e}_c \in \mathcal{V}$, which are stacked into a matrix $\mat{E}$. These vectors are additional parameters in the joint optimization problem \shortmethod\ solves. They can be regarded as views of $C$ typical behaviors around which the trajectories' views are regrouped. To project trajectory $i$'s $\vec{v}_i$ views onto behavioral centroids learned from previous trajectory batches, {\shortmethod} computes a vector of soft assignments of $\vec{v}_i$ to each centroid $\vec{e}_c$:
\begin{equation}
    \mat{Q}_{i}=\underset{ 1 \leq c \leq C}{\text{Softmax}}\bigg(\frac{\vec{v}_{i}^\top \vec{e}_c}{||\vec{v}_{i}||_2||\vec{e}_c||_2}\bigg) 
    \label{eq:Q}
\end{equation}

and forms a $B\times C$ matrix $\mat{Q}$ whose $i$-th row is the soft assignment of $\vec{v}_i$. The resulting assignments may be very unbalanced, with most probability mass assigned to only a few clusters. Applying the Sinkhorn-Knopp algorithm solves this issue by iteratively re-normalizing $\mat{Q}$ in order to obtain a more equal cluster membership~\citep{cuturi2013sinkhorn}, where the degree of re-normalization is controlled by a temperature parameter $\beta$. The output of this operation is a matrix $\tilde{\mat{Q}}$.

For each view $\vec{w}_i$ (the projection of trajectory view $\vec{v}_i$) \shortmethod\ computes the logarithm of its soft cluster assignments and treats these vectors as rows of another $B\times C$ matrix $\mat{P}$:

\begin{equation}
    \mat{P}_{i}=\log \left[ \underset{ 1 \leq c \leq C}{\text{Softmax}}\big(\vec{w}_{i}^\top \vec{e}_c\big) \right].
    \label{eq:P}
\end{equation}

Finally, we compute the cross entropy between $\tilde{\mat{Q}}$ and $\mat{P}$, which measures their inconsistency. This measure is taken as the clustering loss:

\begin{equation}
\mathcal{L}_{\text{clust}}=\text{CrossEntropy}(\tilde{\mat{Q}},\mat{P}).
    \label{eq:sinkhorn_loss}
\end{equation}

\textbf{Encouraging cross-cluster behavioral similarity.}  Note, however, that the clustering loss in the above step emphasizes the recognition of behavioral similarities \emph{within} clusters. 
This may hurt generalization, as the resulting centroids may not faithfully represent behaviors encountered at test time. Our hypothesis is that encouraging encoder $\phi$ to induce latent-space similarities between trajectories from \emph{different but adjacent} clusters will increase its ability to recognize behaviors in unseen test trajectories.

While there are several ways to encourage cross-cluster representational similarity, using a mechanism similar to MYOW~\citep{azabou2021view} on trajectories drawn from neighboring clusters captures this idea particularly well. Namely, to get the cross-predictive loss $\mathcal{L}_{\text{pred}}$, we sample trajectory view pairs from neighboring clusters and apply the cosine-similarity loss to those pairs.

To implement this idea, we define a measure of cluster proximity via a matrix $\mat{D}$ of cosine similarities between cluster centroids: for clusters $k$ and $l$, $\mat{D}_{kl} = ||\vec{e}_k-\vec{e}_l||^2_2$ ~\citep{grill2020bootstrap}. Recall that in the previous step, the clustering branch computed a matrix $\mat{Q}$ whose rows $\mat{Q}_i$ are soft assignments of trajectory $i$'s view $\vec{v}_i$ to clusters. In this step, we convert these soft assignments to hard ones by associating a trajectory's view $\vec{v}_i$ with cluster $c_i = \text{argmax}_{1 \leq c' \leq C} \tilde{\mat{Q}}_i$ and treating a cluster $c$ as consisting of trajectories with indices in the set $\mathbb{T}_c =\{i \mid c=\text{argmax}_{1 \leq c' \leq C} \tilde{\mat{Q}}_i\}$. To assess how predictive a trajectory embedding $\vec{u}_i$ is of a trajectory embedding $\vec{u}_j$, like in the clustering step we will use two special helper maps, $\psi_{\text{pred}}: \mathcal{U} \to \mathcal{V}$ to obtain a reduced-dimensionality view $\vec{v}' = \psi_{\text{pred}}(\vec{u})$ and $\theta_{\text{pred}}$ to further project $\vec{v}'$ to  $\vec{w}' = \theta_{\text{pred}}(\vec{v}')$.

\shortmethod{} proceeds by repeatedly sampling trajectories, which we call \emph{anchor trajectories}, from the batch, with their associated embeddings $\vec{u}$. For each anchor trajectory $n$, consider $K$ clusters $c_1, \ldots, c_K$ nearest to $n$'s cluster $c_n$, as defined by the indices of $K$ largest values in row $c_n$ of matrix $\mat{D}$ (we exclude $c_n$ itself when determining $c_n$'s nearest clusters). Borrowing ideas from the MYOW approach \cite{azabou2021view}, \shortmethod\ mines a view for $\vec{u}_n$ by randomly choosing a trajectory with embedding $\vec{u}^{(n)}_{c_k}$ from each of the neighboring clusters and computing its view $\vec{v}'_{c_k} = \psi_{\text{pred}}(\vec{u}^{(n)}_{c_k})$. We call the neighbors' views  $\vec{v}'_{c_1}, \ldots, \vec{v}'_{c_K}$ trajectory $n$'s \emph{mined views}. 

For the final operation in this step, \shortmethod\ computes trajectory $n$'s predictive view $\vec{w}'_n = \theta_{\text{pred}}(\psi_{\text{pred}}(\vec{u}_n))$ and measures the distance from it to trajectory $n$'s mined views: 
\begin{equation}
    \mathcal{L}^{(n)}_{\text{pred}}=\sum_{k=1}^K||\vec{w}'_n-\vec{v}'_{c_k}||^2_2
\end{equation}

$N$ regulates the number of anchor trajectories to be sampled, so the total prediction loss is
\begin{equation}
    \mathcal{L}_{\text{pred}}=\sum_{n=1}^N \mathcal{L}^{(n)}_{\text{pred}}
    \label{eq:pred_loss}
\end{equation}
\textbf{Updating encoder $\phi$ using reward guidance without reward signal for representation learning.} Note that \shortmethod's total loss $\mathcal{L}_{\text{\shortmethod}}$ depends on the parameters of encoder $\phi$ as well as of clustering networks $\phi_{\text{clust}}$ and $\theta_{\text{clust}}$, prediction networks $\phi_{\text{clust}}$ and $\theta_{\text{clust}}$, and cluster centroids $\vec{e}_c$, $1 \leq c \leq C$. In each epoch, \shortmethod\ updates all these parameters to minimize $\mathcal{L}_{\text{\shortmethod}}$.

\section{Connection to bisimulation}

Deep bisimulation metrics are tightly connected to the underlying mechanism of mining behaviorally similar trajectories of {\shortmethod}. They operate on a latent-dimensional space and, as is the case for DeepMDP~\citep{gelada2019deepmdp} and DBC~\citep{zhang2020learning}, ensure that bisimilar states (i.e. behaviorally similar states with identical reward) are located close to each other in that latent space. In this section, we aim to highlight a functional similarity between bisimulation metrics and {\shortmethod}.

\begin{definition}
  A bisimilation relation $E\subseteq \mathcal{S}\times \mathcal{S}$ is a binary relation which satisfies,  $\forall(s,t)\in E$:
\begin{enumerate}
    \item $\forall a\in \mathcal{A}, \mathcal{R}(s,a)=\mathcal{R}(t,a)$\;
    \item $\forall a\in \mathcal{A}, \forall c\in \mathcal{S},\sum_{s'\in c} \mathcal{P}(s,a)(s')=\sum_{s'\in c} \mathcal{P}(t,a)(s')$
\end{enumerate}
\end{definition}

In practice, rewards and transition probabilities rarely match exactly. For this reason, \cite{ferns2004metrics} proposed a smooth alternative to bisimulation relations in the form of bisimulation metrics, which can be found by solving a recursive equation involving the Wasserstein-1 distance $\mathcal{W}_1$ between transition probabilities. $W_1$ can be found by solving the following linear programming~\citep{villani2008optimal}, where we let $\Gamma=\{\vec{v}\in \mathbb{R}^{|\mathcal{V}|}:0\leq  \vec{v}_i\leq 1 \; \forall 1 \leq i \leq |\mathcal{V}|\}$:
\begin{equation}
        \mathcal{W}_1^d(P||Q):=\max_{\mu\in \Gamma} \sum_{s\in \mathcal{S}}(P(s)-Q(s))\mu(s)
     \quad \text{s.t.} \;\mu(s)-\mu(s')<d(s,s') \forall s,s' \in \mathcal{S},
    \label{eq:bisimulation_mat_form}
\end{equation}
where $\mu$ is a vector whose elements are constrained between 0 and 1. In practice, bisimulation metrics are used to enforce a temporal continuity of the latent space by minimization of the $\mathcal{W}_1$ loss between training state-action pairs. Therefore, to show a connection of {\shortmethod} to (reward-free) bisimulation metrics, it is sufficient to show that two trajectories are mapped to the same partition \emph{if} their induced $\mathcal{W}_1$ distance is arbitrarily small. In our (informal) argument that follows, we assume that {\shortmethod} samples two consecutive timesteps and encodes them into $\vec{v}$; the exact form of $\vec{v}$ dictates the nature of the behavioral similarity. The proof can be found in Appendix ~\ref{app:proofs}.

\begin{proposition}(Informal)
Let $M$ be an MDP where $\mathcal{R}(s,a)=0$ for all $(s,a)\in \mathcal{S}\times \mathcal{A}$ and let $\vec{v},\vec{v}'\in\mathcal{V}$ be two dynamics embeddings in $M$. The clustering operation between $\vec{v},\vec{v}'$ induces a reward-free bisimilarity metric $\mathcal{W}_1(\mathbb{P}[\vec{v}],\mathbb{P}[\vec{v}'])$ between induced distributions $\mathbb{P}[\vec{v}]$ and $\mathbb{P}[\vec{v}']$.
\label{thm:bisimulation_partition}
\end{proposition}

\section{Empirical evaluation}\label{sec:results}
We compare \shortmethod{} against strong RL baselines: DAAC~\citep{raileanu2021decoupling} -- the current state-of-the-art on the challenging generalization benchmark suite \emph{Procgen}~\citep{cobbe2020leveraging}, and PPO~\citep{schulman2017proximal}.
DAAC optimizes the PPO loss \citep{schulman2017proximal} through decoupling the training of the policy and value functions, which updates the advantage function during the policy network updates. 
We then compare to several unsupervised and SSL auxiliary objectives used in conjunction with PPO.
DIAYN~\citep{eysenbach2018diversity} is an unsupervised skill-based exploration method which we adapt to the online setting by uniformly sampling skills. Its notion of skills has some similarities to the notion of clusters in \shortmethod.
We also compare with two SSL-based auxiliary objectives: CURL~\citep{srinivas2020curl}, a common SSL baseline which contrasts augmented instances of the same state, and Proto-RL~\citep{yarats2021reinforcement}, which we adapt for this generalization setting. 
Finally, we provide a comparison against bisimulation-based algorithms: DBC~\citep{zhang2020learning}, which was shown to perform well on robotic control tasks with visual distractor features, and PSE~\citep{agarwal2021contrastive}. PSE assumes policies for training tasks to be given and, like \citet{agarwal2021contrastive}, we ran it both with random and high-quality policies pretrained with extra computation budget. See Appendix \ref{sec:experiment_details} for details.

The Procgen benchmark suite, which we use in our experiments, consists of 16 video games (see Table~\ref{tab:procgen}). Procgen procedurally generates distinct levels for each game. The number of levels for each game is virtually unlimited. Levels within a game which share common game rules and objectives but differ in level design such as the number of projectiles, background colors, item placements throughout the level and other game assets. All of this makes Procgen a suitable benchmark for zero-shot generalization. Using our notation from \Cref{sec:problem}, for each of 16 games, we train on a uniform distribution $d(\mathscr{T}_N)$ over $N = 200$ ``easy'' levels of the game and evaluate on $d(\mathscr{T}\setminus \mathscr{T}_N)$, i.e., a uniform distribution over the game's ``easy'' levels not seen during training. Following \cite{mohanty2021measuring}, we report results after 8M steps of training, since this demonstrates the quality of ZSG that various representation learning methods can achieve \emph{quickly}. However, \Cref{fig:eval_curves_25M} in Appendix \ref{sec:addit} also provides results after 25M steps of training, as in the original Procgen paper \citep{cobbe2020leveraging}.

\begin{table}[ht]
\centering
\resizebox{\linewidth}{!}{%
\begin{tabular}{l||cc|ccc|ccc|c}
\toprule
& \multicolumn{2}{c|}{RL} &  \multicolumn{3}{c|}{RL+Bisim.} & RL+Unsup. & \multicolumn{2}{c|}{RL+SSL} & Ours \\
\midrule 
Env & PPO & DAAC & PPO+DBC & PPO+PSE (random) & PPO+PSE (pretrained) & PPO+DIAYN & Proto-RL & PPO+CURL & \textbf{{\shortmethod}} \\
\midrule
bigfish & 2.3$\pm$0.1 & 4.3$\pm$0.3 & 1.8$\pm$0.1 & 2.3 $\pm$ 0.1 & 1.8 $\pm$ 0.2  & 2.2$\pm$0.1 & 2.4$\pm$0.1 & 2.2$\pm$0.2 & \textbf{4.7$\pm$0.2} \\
bossfight & 5.2$\pm$0.3 & 1.7$\pm$0.7 & 5$\pm$0.1 & 0.9 $\pm$ 0.2 & 0.7 $\pm$ 0.1 & 1.1$\pm$0.2 & 6.1$\pm$0.5 & 4.6$\pm$0.8 & \textbf{8.2$\pm$0.1} \\
caveflyer & 4.4$\pm$0.3 & 4.3$\pm$0.1 & 3.6$\pm$0.1 & 2.6 $\pm$ 0.1 & 3.6 $\pm$ 0.3 & 1.0$\pm$1.9 & \textbf{4.7$\pm$0.1} & 4.6$\pm$0.3  & \textbf{4.7$\pm$0.2} \\
chaser & 7.2$\pm$0.2 & 7.1$\pm$0.1 & 4.8$\pm$0.1 & \textbf{8.7 $\pm$ 0.5} & 4.2 $\pm$ 0.2 & 5.6$\pm$0.5 & 7.6$\pm$0.2 & 7.2$\pm$0.2  & 7.1$\pm$0.2 \\
climber & 5.1$\pm$0.1 & 5.5$\pm$0.2 & 4.1$\pm$0.4  & 2.9 $\pm$ 0.1 & 3.9 $\pm$ 0.2 & 0.8$\pm$0.7 & 5.5$\pm$0.3 & 5.5$\pm$0.1 & \textbf{5.9$\pm$0.2} \\
coinrun & 8.3$\pm$0.2 & 8.1$\pm$0.1 & 7.9$\pm$0.1 & 5.5 $\pm$ 0.5 & 7.3 $\pm$ 0.2 & 6.4$\pm$2.6 & 8.2$\pm$0.1 & 8.1$\pm$0.1  & \textbf{8.7$\pm$0.3} \\
dodgeball & 1.3$\pm$0.1 & \textbf{1.8$\pm$0.2} & 1.0$\pm$0.3 & 1.6 $\pm$ 0.1 & 1.3 $\pm$ 0.1 & 1.4$\pm$0.2 & 1.6$\pm$0.1 & 1.4$\pm$0.1 & \textbf{1.8$\pm$0.1} \\
fruitbot & 12.4$\pm$0.2 & 11.5$\pm$0.3 & 7.6$\pm$0.2 & 1.0 $\pm$ 0.1 & 1.1 $\pm$ 0.2 & 7.2$\pm$3.0 & 12.3$\pm$0.4 & 12.3$\pm$0.2  & \textbf{13.3$\pm$0.3} \\
heist & 2.7$\pm$0.2 & \textbf{3.4$\pm$0.2} & 3.3$\pm$0.2 & 3.1 $\pm$ 0.3 & 2.9 $\pm$ 0.4 & 0.2$\pm$0.2 & 3.0$\pm$0.3 & 2.5$\pm$0.1  & 3.1$\pm$0.3 \\
jumper & 5.8$\pm$0.3 & \textbf{6.3$\pm$0.1} & 3.9$\pm$0.4 & 4.1 $\pm$ 0.3 & 5.3 $\pm$ 0.2 & 2.6$\pm$2.3 & 6.0$\pm$0.1 & 5.9$\pm$0.1  & 6.0$\pm$0.1 \\
leaper & 3.5$\pm$0.4 & 3.5$\pm$0.4 & 2.7$\pm$0.1 & 2.7 $\pm$ 0.2 & 2.6 $\pm$ 0.1 & 2.5$\pm$0.2 & 3.2$\pm$0.8 & \textbf{3.6$\pm$0.5}  & 2.8$\pm$0.2 \\
maze & 5.4$\pm$0.2 & 5.6$\pm$0.2 & 5.0$\pm$0.1 & 5.4 $\pm$ 0.2 & 5.3 $\pm$ 0.1 & 1.6$\pm$1.2 & 5.5$\pm$0.3 & 5.4$\pm$0.1 & \textbf{5.7$\pm$0.1} \\
miner & 8.7$\pm$0.3 & 5.7$\pm$0.1 & 4.8$\pm$0.1 & 5.6 $\pm$ 0.1 & 4.3 $\pm$ 0.3 & 1.3$\pm$2.0 & 8.8$\pm$0.5 & \textbf{8.6$\pm$0.2}  & 6.5$\pm$0.2 \\
ninja & 5.5$\pm$0.2 & 5.2$\pm$0.1 & 3.5$\pm$0.1 & 3.4 $\pm$ 0.2 & 3.5 $\pm$ 0.3 & 2.8$\pm$2.0 & 5.4$\pm$0.3 & 5.6$\pm$0.1  & \textbf{5.8$\pm$0.1} \\
plunder & 6.2$\pm$0.4 & 4.1$\pm$0.1 & 5.1$\pm$0.1 & 4.0 $\pm$ 0.1 & 4.1 $\pm$ 0.2 & 2.1$\pm$2.5 & 6.0$\pm$0.9 & 6.5$\pm$0.3 & \textbf{6.6$\pm$0.3} \\
starpilot & 4.7$\pm$0.2 & 4.1$\pm$0.2 & 2.8$\pm$0.1 & 3.2 $\pm$ 0.2 & 3.0 $\pm$ 0.1 & 5.8$\pm$0.6 & 5.2$\pm$0.2 & 5.0$\pm$0.1 & \textbf{7.7$\pm$0.5} \\
\bottomrule
\end{tabular}%
}
\caption{\small Average evaluation returns collected after 8M training frames, $\pm$ one standard deviation over 10 seeds.} 
\label{tab:procgen}
\end{table}%

\textbf{Main results.} As Table~\ref{tab:procgen} shows, PPO+\shortmethod\ outperforms all other baselines, including  DAAC, on most games. Notably, bisimulation-based approaches other than \shortmethod -- DBC as well as PSE with both random and expert data-gathering policies --   exhibit lower gains than others. While this can be surprising, recent work has seen similar results when applying DBC to tasks with unseen backgrounds~\citep{li2021domain}. PSE's inferior performance may be due to the policy similarity metric, which PSE requires as input and which we took from \citet{agarwal2021contrastive}, being poorly suited to Procgen. This highlights an important difference between \shortmethod\ and PSE: \shortmethod\ doesn’t need a policy similarity metric, since it implicitly induces such a metric based on trajectory “signatures”. Despite training static prototypes for 8M timesteps and adapting the RL head for 8M additional ones (see Appendix \ref{sec:experiment_details} for details), Proto-RL performs worse than {\shortmethod}. This suggests that the temporal aspect of clustering is key for ZSG, a hypothesis we explore further in \Cref{sec:slow}. Likewise, PPO+DIAYN uses its pre-training phase to find a diverse set of skills, which can be useful in robotics domains, but does not help much in the ZSG setting of Procgen. DAAC also exhibits good generalization performance, but inherits from PPG~\citep{cobbe2020phasic} the separation of value and policy functions, an overhead which {\shortmethod} manages to avoid. In addition, we describe a number of ablation studies (Appendix \ref{sec:addit}), empirically show that slow clustering convergence leads to better generalization (Appendix \ref{sec:slow}), and demonstrate on a toy task how learning behavioral similarities captures local changes (Appendix \ref{sec:behsim}).

\section{Conclusions}\label{sec:discussion}
This work proposed {\shortmethod}, a novel representation learning algorithm that facilitates zero-shot generalization of RL policies in high-dimensional observation spaces. \shortmethod\ can be viewed as inducing an unsupervised reward-agnostic bisimulation metric over observation histories, learned over transitions encountered by policies from an RL algorithm's \emph{value improvement path}~\citep{dabney2021valueimprovement}. We hope that in the future \shortmethod\ will inspire other representation learning methods based on capturing belief states's behavioral similarity, which will be capable of policy generalization across greater variations in environment dynamics.

\newpage
\section{Appendix}

\subsection{\shortmethod\ pseudocode \label{sec:pseudocode}}

\begin{algorithm}[H]
 \SetAlgoLined
 \SetKwInOut{Inputs}{Inputs}
 \SetKwInOut{Hyperparameters}{Hyperparameters}
 \SetKwInOut{Output}{Output}
 \Inputs{online encoder $\phi$, cluster projector  $\theta_{\text{clust}}$, cluster encoder $\psi_{\text{clust}}$, mining projector $\theta_{\text{pred}}$, mining encoder $\psi_{\text{pred}}$ , cluster basis matrix $\mat{E}$}
 \Hyperparameters{$B$ -- trajectory batch size, $C$ -- num. of trajectory clusters, $T$ -- subtrajectory length, $K$ -- num. of nearest clusters for view mining, $L$ -- trajectory length, $N$ -- num. of anchors for view mining, $\beta$ -- Sinkhorn temperature}
 \For{each iteration $itr=1,2,..$}{
 \For{each minibatch $\mathcal{B}$}{

\For{each trajectory $\tau_i$ in $\mathcal{B}$}{
$t_1,..,t_T\sim $Uniform($L$) \ \tcp{Sample temporal keypoints}

$u^{(\tau_i)}=[FiLM(\phi(s_{t_1}),a_{t_1}),...,FiLM(\phi(s_{t_T}),a_{t_T})]$ \
}

\tcp{cluster dynamics}
$\vec{u}_i = [u^{(\tau_1)},...,u^{(\tau_m)}]$ \ \tcp{ batch dynamics}

$\vec{v}_{i}=\psi_{\text{clust}}(\vec{u}_i)$ \ \tcp{fetch embeddings}

$\vec{w}_{i}=\theta_{\text{clust}}(\vec{v}_{i})$ \ \tcp{fetch projections}

$\vec{v}_{i} = \frac{\vec{v}_{i}}{||\vec{v}_{i}||_2}$ \tcp{normalize embeddings}

$\mat{Q} = \text{Softmax}\big(\vec{v}_{i}^\top \mat{E}  / \beta\big)$ \ \tcp{Compute latent dynamics scores}

$\tilde{\mat{Q}} = \text{Sinkhorn}(\mat{Q}) $ \ \tcp{normalize scores through Sinkhorn}

$\mat{P} =  \log \left[\text{Softmax} (\vec{w}_{i}^\top \mat{E} / \beta) \right]$ \ \tcp{Compute projected dynamics scores}

$\mathcal{L}_{\text{clust}}(\phi,\psi_{\text{cluster}},\theta_{\text{cluster}})=\text{CrossEntropy}(\tilde{\mat{Q}},\mat{P})$\

\tcp{Predicting neighbors}
$\mat{D}_{ij} = ||\vec{e}_i-\vec{e}_j||^2_2$ \tcp{find pairwise basis distances}
$\mathcal{L}_{\text{pred}} = 0$

\For{each anchor $j = 1,..., N$}{
$\tau_j \sim \mathcal{B}$ \tcp{Sample anchor trajectory}
$\vec{u}_n = u^{(\tau_j)}$ \tcp{Set anchor embedding}
$c_i^{(1)},..,c_i^{(k)}=\text{top-knn}(\mat{D},k,c_i)$ \tcp{Find nearby clusters}
$u_{c_1}, ..., u_{c_k} \sim p(u_{c_i^{(1)}}), ..., p(u_{c_i^{(k)}})$ \tcp{Sample views from clusters}

$\vec{v}'_{c_1}, \ldots, \vec{v}'_{c_K} = \psi_{\text{pred}}(u_{c_1}), ..., \psi_{\text{pred}}(u_{c_k})$ \tcp{embed mined views} \

$\vec{w}'_n = \theta_{\text{pred}}(\psi_{\text{pred}}(\vec{u}_n))$ 
\tcp{mining target}

$\mathcal{L}^{(n)}_{\text{pred}} = \sum_{k=1}^K||\vec{w}'_n-\text{StopGrad}(v'_{c_k})||^2_2$
}

$\mathcal{L}_{\text{pred}}=\sum_{n=1}^N \mathcal{L}^{(n)}_{\text{pred}}$

$\mathcal{L}_{\text{\shortmethod}} = \mathcal{L}_{\text{clust}} + \mathcal{L}_{\text{pred}}$ \tcp{update networks}

$\phi,\psi_{\text{clust}},\theta_{\text{clust}}, \theta_{\text{pred}}, \psi_{\text{pred}} = \text{Adam}(\phi,\psi_{\text{clust}},\theta_{\text{clust}}, \theta_{\text{pred}}, \psi_{\text{pred}} ; \mathcal{L}_{\text{\shortmethod}})$ \ 
 }
 
 \For{each minibatch $\mathcal{B}$}{

$\pi,V^\pi = \text{Adam}(\pi,V^\pi; \mathcal{L}_{\text{RL}}(\mathcal{B}))$ \tcp{update RL parameters}

 }
 }
  \caption{\longmethod}
  \label{alg:main}
 
 \end{algorithm}

\comment {
\subsection{Detailed algorithm description \label{sec:details}}

\shortmethod's pseudocode is presented in \Cref{alg:main} and the schema in \Cref{fig:ctrl} in the main paper. Below we describe the details of each of \shortmethod's steps. While in general the agent's belief state at step $t$ of a trajectory is the entire observation history $h_t = (o_1, \ldots, o_t)$, in the rest of the section we will assume $h_t = (o_t)$ and, in a slight abuse of notation, use $\phi(o_t)$ instead of $\phi(h_t)$ to simplify explanations\footnote{While, in theory, the Procgen suite is indeed a POMDP, most RL algorithms take the most recent observation as the belief state -- a simplification which was shown not to hinder ZSG on Procgen~\citep{cobbe2020leveraging}.}. We emphasize, however, that \shortmethod\ equally applies in settings where the agent uses a much longer history of observations as its state.

\comment{
Like most online RL methods themselves, our algorithm operates in epochs, learning from a batch of trajectories in each epoch. Before the RL algorithm updates the policy head $\theta$ over the epoch's trajectory batch, \shortmethod\ uses this trajectory batch to update the encoder $\phi$ with 4 operations:

\begin{enumerate}
    \item Apply $\phi$ and additional projections to stochastically generate a low-dimensional \emph{reward-agnostic} view of each trajectory from the batch.

    \item Group the trajectory views with the clustering procedure into $C$ partitions, where $C$ is \shortmethod's hyperparameter.
    Compute a clustering loss $\mathcal{L}_{\text{clust}}$.
    
    \item Sample pairs of trajectories whose views are in nearby clusters. For each trajectory in a pair, construct its mined view. Compute a prediction loss $\mathcal{L}_{pred}$ over the mined views of the chosen trajectory pairs.
    
    \item Update $\phi$, cluster centroids, and other parameters involved in computing the views w.r.t. the \shortmethod\ objective $\mathcal{L}_{\text{\shortmethod}} = \mathcal{L}_{\text{clust}} + \mathcal{L}_{\text{pred}}$.
\end{enumerate}

We refer the reader to \shortmethod's pseudocode in \Cref{alg:main} (Appendix) and the schema in \Cref{fig:ctrl}. While in general the agent's belief state at step $t$ of a trajectory is the entire observation history $h_t = (o_1, \ldots, o_t)$, in the rest of the section we will assume $h_t = (o_t)$ and, in a slight abuse of notation, use $\phi(o_t)$ instead of $\phi(h_t)$ to simplify explanations\footnote{While, in theory, the Procgen suite is indeed a POMDP, most practical applications treat it's belief state as being the most recent observation -- a simplification which was shown not to hinder ZSG~\citep{cobbe2020leveraging}.}. We emphasize, however, that \shortmethod\ equally applies in settings where the agent uses the entire history of observations as its state.
}

\textbf{Generating low-dimensional trajectory views with encoder $\phi$.} \shortmethod's input in each epoch is a trajectory batch $\{traj_i\}_{i=1}^B$ of size $B$. Assume all trajectories in the batch have the same length $L$. For an integer hyperparameter $T \leq L$, for each trajectory $traj_i=(o_0, a_0, r_0, \ldots o_{T_b}, a_{L},r_{L})$ we independently and uniformly sample a subset of its steps $\tau_i = t_1, \ldots, t_T$ to form a subtrajectory $(o_{t_1}, a_{t_1}, r_{t_1}, \ldots o_{t_T}, a_{t_T},r_{t_T})$. We then encode this subtrajectory as 
\begin{equation}
\vec{u}_i^{(\tau_i)}=(FiLM(\phi(o_{t_1}),a_{t_1}),..,FiLM(\phi(o_{t_T}),a_{t_T})),
    \label{eq:clustering_tuples}
\end{equation}

where $FiLM(\phi(o_{t_j}),a_{t_j})$ is a common way of combining representations of different objects, akin to conditioning \cite{perez2017film}, and the resulting vector $\vec{u}_i^{(\tau_i)}$ is in a low-dimensional space $\mathcal{U}$. Note two aspects of the process of generating these vectors: (1) it drops rewards from the original trajectory and (2) it critically relies on $\phi$ whose parameter values are learned from previous epochs. Vectors $\vec{u}_i^{(\tau_i)}$ produced in this way are the trajectory views that the next steps of \shortmethod\ operate on.\\

\textbf{Clustering trajectory views.} \shortmethod\ groups trajectories from the epoch's batch by clustering the set of their views $\{\vec{u}_i^{(\tau_i)}\}_{i=1}^B$ using the online Sinkhorn-Knopp procedure~\citep{caron2020unsupervised}. Since \shortmethod\ is an online algorithm, Sinkhorn-Knopp is better-suited for the task than other clustering methods.

The clustering branch computes two views of each input $\vec{u}_i^{(\tau_i)}$ in a cascading fashion: first by passing it through a clustering encoder $\psi_{\text{clust}}:\mathcal{U} \to \mathcal{V}$, e.g. an RNN, to obtain a lower-dimensional view  $\vec{v}_i=\psi_{\text{clust}}(\vec{u}_i^{(\tau_i)})$, and then by passing $\vec{v}_i$ through yet another network, an MLP $\theta_{\text{clust}}:\mathcal{V}\to \mathcal{W}$, to produce view $\vec{w}_i=\theta_{\text{clust}}(\vec{v}_i)$. The parameters of $\psi_{\text{clust}}$ and $\theta_{\text{clust}}$ are learned through the epochs jointly with $\phi$'s. Like $\vec{u}_i^{(\tau_i)}$, each $\vec{v}_i$ and $\vec{w}_i$ is a view of trajectory $i$; the approach then consists in projecting $\vec{v}_i$'s and $\vec{w}_i$'s onto centroids of $C$ clusters in two different ways and then computes a clustering loss that enforces consistency between $\vec{v}_i$'s and $\vec{w}_i$'s cluster projections.

Specifically, we represent the centroid of each cluster $c$ with a vector $\vec{e}_c \in \mathcal{V}$, which are stacked into a matrix $\mat{E}$. These vectors are additional parameters in the joint optimization problem \shortmethod\ solves. They can be regarded as views of $C$ typical behaviors around which the trajectories' views are regrouped. To project trajectory $i$'s $\vec{v}_i$ views onto behavioral centroids learned from previous trajectory batches, {\shortmethod} computes a vector of soft assignments of $\vec{v}_i$ to each centroid $\vec{e}_c$:
\begin{equation}
    \mat{Q}_{i}=\underset{ 1 \leq c \leq C}{\text{Softmax}}\bigg(\frac{\vec{v}_{i}^\top \vec{e}_c}{||\vec{v}_{i}||_2||\vec{e}_c||_2}\bigg) 
    \label{eq:Q}
\end{equation}

and forms a $B\times C$ matrix $\mat{Q}$ whose $i$-th row is the soft assignment of $\vec{v}_i$. The resulting assignments may be very unbalanced, with most probability mass assigned to only a few clusters. Applying the Sinkhorn-Knopp algorithm solves this issue by iteratively re-normalizing $\mat{Q}$ in order to obtain a more equal cluster membership~\citep{cuturi2013sinkhorn}, where the degree of re-normalization is controlled by a temperature parameter $\beta$. The output of this operation is a matrix $\tilde{\mat{Q}}$.

For each view $\vec{w}_i$, the projection of trajectory view $\vec{v}_i$, \shortmethod\ computes the logarithm of its soft cluster assignments:

\begin{equation}
    \mat{P}_{i}=\log \left[ \underset{ 1 \leq c \leq C}{\text{Softmax}}\big(\vec{w}_{i}^\top \vec{e}_c\big) \right],
    \label{eq:P}
\end{equation}

and treats these vectors as rows of another $B\times C$ matrix $\mat{P}$.

Finally, we compute the cross entropy between $\tilde{\mat{Q}}$ and $\mat{P}$, which measures their inconsistency. This measure is taken as the clustering loss:

\begin{equation}
\mathcal{L}_{\text{clust}}=\text{CrossEntropy}(\tilde{\mat{Q}},\mat{P})
    \label{eq:sinkhorn_loss}
\end{equation}

\textbf{Encouraging cross-cluster behavioral similarity.} To implement this idea, we define a measure of cluster proximity via a matrix $\mat{D}$ of cosine similarities between cluster centroids: for clusters $k$ and $l$, $\mat{D}_{kl} = ||\vec{e}_k-\vec{e}_l||^2_2$ ~\citep{grill2020bootstrap}. Recall that in the previous step, the clustering branch computed a matrix $\mat{Q}$ whose rows $\mat{Q}_i$ are soft assignments of trajectory $i$'s view $\vec{v}_i$ to clusters. In this step, we convert these soft assignments to hard ones by associating a trajectory's view $\vec{v}_i$ with cluster $c_i = \text{argmax}_{1 \leq c' \leq C} \tilde{\mat{Q}}_i$ and treating a cluster $c$ as consisting of trajectories with indices in the set $\mathbb{T}_c =\{i \mid c=\text{argmax}_{1 \leq c' \leq C} \tilde{\mat{Q}}_i\}$. To assess how predictive a trajectory embedding $\vec{u}_i$ is of a trajectory embedding $\vec{u}_j$, like in the clustering step we will use two special helper maps, $\psi_{\text{pred}}: \mathcal{U} \to \mathcal{V}$ to obtain a reduced-dimensionality view $\vec{v}' = \psi_{\text{pred}}(\vec{u})$ and $\theta_{\text{pred}}$ to further project $\vec{v}'$ to  $\vec{w}' = \theta_{\text{pred}}(\vec{v}')$.

\shortmethod{} proceeds by repeatedly sampling trajectories, which we call \emph{anchor trajectories}, from the batch, with their associated embeddings $\vec{u}$. For each anchor trajectory $n$, consider $K$ clusters $c_1, \ldots, c_K$ nearest to $n$'s cluster $c_n$, as defined by the indices of $K$ largest values in row $c_n$ of matrix $\mat{D}$ (we exclude $c_n$ itself when determining $c_n$'s nearest clusters). Borrowing ideas from the MYOW approach \cite{azabou2021view}, \shortmethod\ mines a view for $\vec{u}_n$ by randomly choosing a trajectory with embedding $\vec{u}^{(n)}_{c_k}$ from each of the neighboring clusters and computing its view $\vec{v}'_{c_k} = \psi_{\text{pred}}(\vec{u}^{(n)}_{c_k})$. We call the neighbors' views  $\vec{v}'_{c_1}, \ldots, \vec{v}'_{c_K}$ trajectory $n$'s \emph{mined views}. 

For the final operation in this step, \shortmethod\ computes trajectory $n$'s predictive view $\vec{w}'_n = \theta_{\text{pred}}(\psi_{\text{pred}}(\vec{u}_n))$ and measures the distance from it to trajectory $n$'s mined views: 
\begin{equation}
    \mathcal{L}^{(n)}_{\text{pred}}=\sum_{k=1}^K||\vec{w}'_n-\vec{v}'_{c_k}||^2_2
\end{equation}

$N$ is the \shortmethod's hyperparameter that determines the number of anchor trajectories to be sampled, so the total prediction loss is
\begin{equation}
    \mathcal{L}_{\text{pred}}=\sum_{n=1}^N \mathcal{L}^{(n)}_{\text{pred}}
\end{equation}

\ \\
\textbf{Updating encoder $\phi$.} \shortmethod's loss is a combination of the clustering and prediction losses: 
\begin{equation}
\mathcal{L}_{\text{\shortmethod}} = \mathcal{L}_{\text{clust}} + \mathcal{L}_{\text{pred}}
\end{equation}
Note that this loss depends on the parameters of encoder $\phi$ as well as of clustering networks $\phi_{\text{clust}}$ and $\theta_{\text{clust}}$, prediction networks $\phi_{\text{clust}}$ and $\theta_{\text{clust}}$, and cluster centroids $\vec{e}_c$, $1 \leq c \leq C$. In each epoch, \shortmethod\ updates all these parameters to minimize $\mathcal{L}_{\text{\shortmethod}}$.
}

\subsection{Experiment details}
\label{sec:experiment_details}
\begin{table}[h]
    \centering
    \begin{tabular}{c|l|c}
        Name & Description & Value \\
        \hline 
       $\gamma$ & Discount factor & 0.999\\
       $\lambda$ & Decay & 0.95\\
       $n_\text{timesteps}$ & Number of timesteps per rollout & 256\\
       $n_\text{epochs}$ & Number of epochs for RL and representation learning & 1\\
       $n_\text{samples}$ & Number of samples per epoch & 8192\\
       Entropy bonus & Entropy loss coefficient & 0.01\\
       Clip range & Clip range for PPO & 0.2\\
       Learning rate & Learning rate for RL and representation learning & $5\times 10^{-4}$\\
       Number of environments & Number of parallel environments & 32\\
       Optimizer & Optimizer for RL and representation learning & Adam\\
       Frame stack & Frame stack $X$ Procgen frames & 1\\
       \midrule
       $E$ & Number of clusters & 200\\
       $k$ & Number of k-NN nearest neighbors & 3\\
       $T$ & Number of clustering timesteps & 2\\
       $\beta$ & Clustering temperature & 0.3
    \end{tabular}
    \caption{\small Experiments' parameters}
    \label{tab:appendix_experiment_params}
\end{table}

We implemented all algorithms on top of the IMPALA architecture, which was shown to perform well on Procgen~\citep{cobbe2020leveraging}. In the Procgen experiments, \textbf{Proto-RL} was ran without any intrinsic rewards (since the domains are not exploration-focused) by first jointly training the representation and RL losses for 8M timesteps, after which only the RL loss was optimized for an additional 8M steps (16M steps in total). Similarly, \textbf{DIAYN} was also run with the pre-training phase of 8M and then RL only objective for the second 8M phase.

Like Proto-RL and DIAYN, \textbf{PSE} needed extra training budget and extra adjustments for a fair comparison to \shortmethod\ (see \Cref{sec:rel_work}). \citet{agarwal2021contrastive} used PSE only with Soft Actor-Critic, which doesn’t perform well on Procgen. Therefore, we carried over PSE' available implementation from  \href{https://agarwl.github.io/pse/}{https://agarwl.github.io/pse/} to our codebase, with the help of PSE' authors, to combine it with the same PPO implementation that \shortmethod\ used. 

PSE assumes being given policies for training problem instances (Procgen levels). Like \citet{agarwal2021contrastive}, we ran PSE using both random and high-quality pretrained policies for these problems. In the latter case, we pretrained expert policies for the first 40 levels of Procgen to generate training trajectories for PSE. Each level’s expert was trained on 0.5M environment steps. Note that this is much less than 8M steps we used for policy training in experiments with other algorithms, but this is because each expert needed to be good \emph{only for a single level}, and we verified that they indeed were. We did this only for the first 40 levels because even for 40 levels this took 20M training steps \emph{per game} and had to be done for 16 games. We don’t believe more than 40 experts per game would have made a difference.

Given policies for the training levels, PPO+PSE's training on Procgen mimicked \shortmethod’s: PPO+PSE trained by interacting with the first 200 levels for 8M steps and was evaluated on the rest. However, during the training, PPO+PSE sampled an additional 1M interactions from the pretrained traiing-level policies. the process was repeated for all 16 games, for 10 seeds each.

We did hyperparameter grid search on PPO+PSE’s hyperparameters for PSE -- loss coefficient values of $(0.1,1,2)$ and temperature $(0.1,0.3,0.7)$. PPO’s hyperparameters were the same as in \shortmethod.

Thus, due to the need to pretrain and gather data with per-level expert policies, PPO+PSE received  $0.5 \cdot 20M + 1M = 21M$ extra environment interactions compared to \shortmethod, i.e., used $21M/8M = 2.6\times$ more training data than the latter.

\subsection{Additional results \label{sec:addit}}

\paragraph{ZSG over 8M and 25M training steps and policy performance throughout training}

\begin{figure}[h!]
    \centering
    \includegraphics[width=0.9\linewidth]{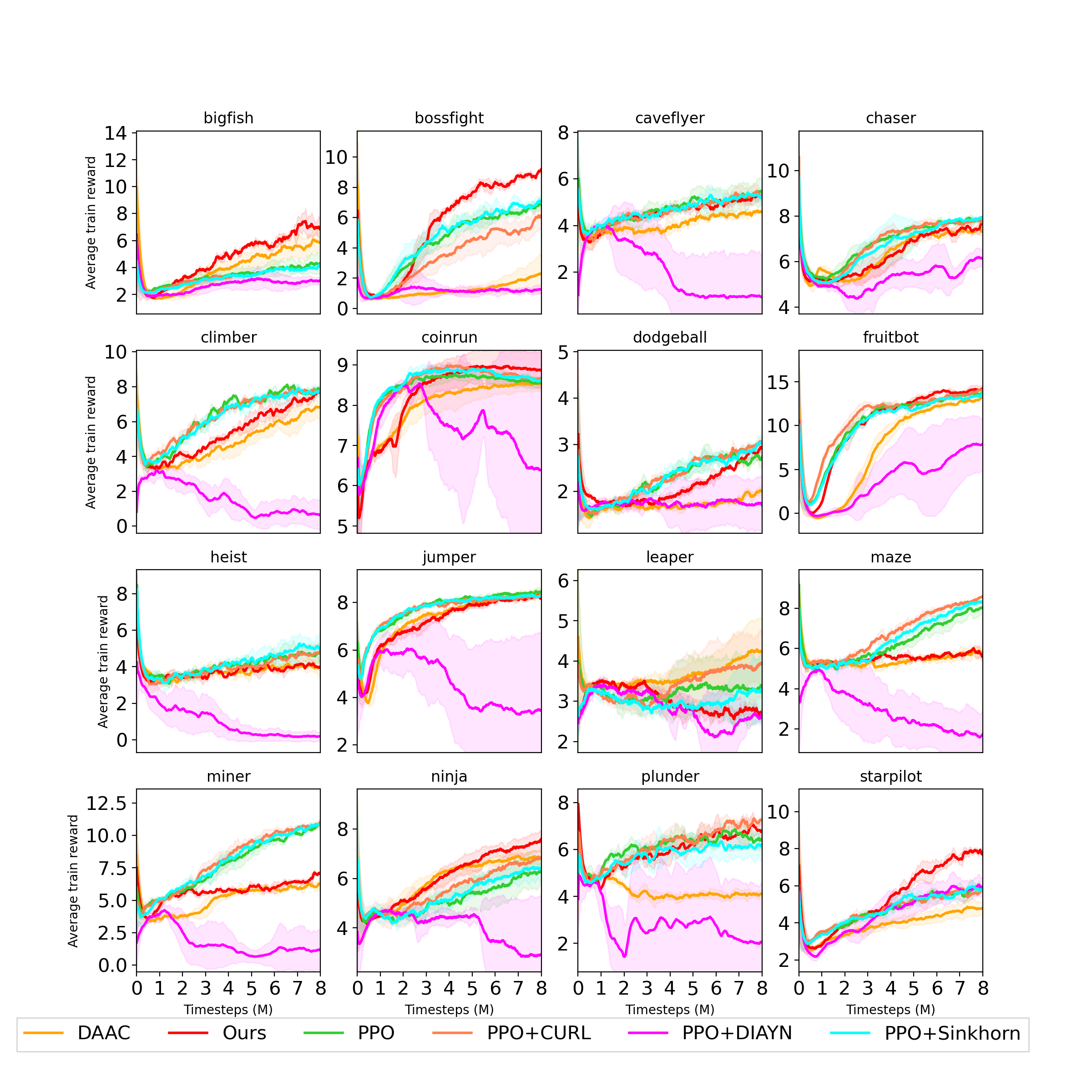}
    \caption{\small Training results over the 8M frames benchmark.}
    \label{fig:train_curves_8M}
\end{figure}

\begin{figure}[h!]
    \centering
    \includegraphics[width=0.9\linewidth]{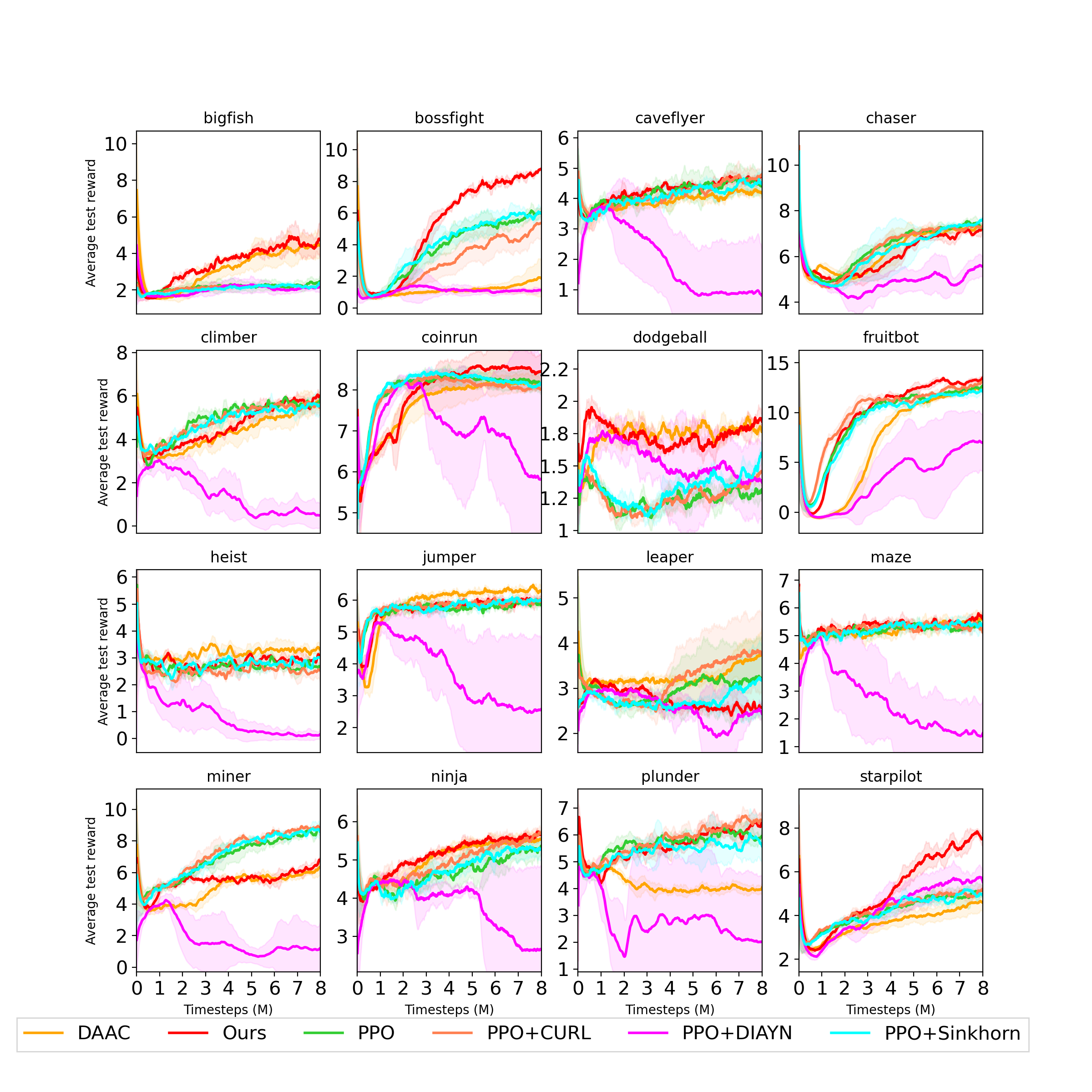}
    \caption{\small Evaluation results over the 8M frames benchmark.}
    \label{fig:eval_curves_8M}
\end{figure}

\begin{figure}[h!]
    \centering
    \includegraphics[width=0.9\linewidth]{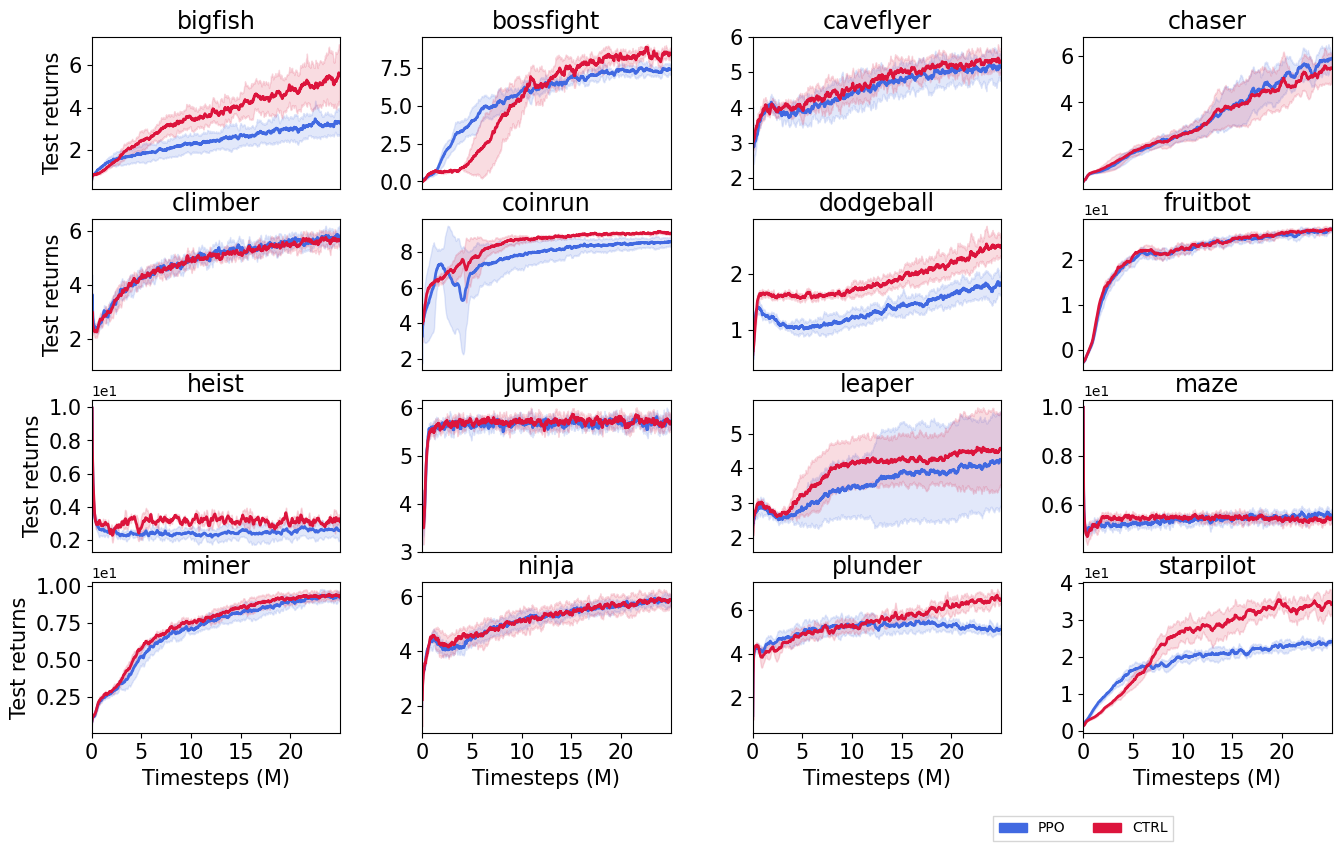}
    \caption{\small Evaluation results over the 25M frames benchmark.}
    \label{fig:eval_curves_25M}
\end{figure}

We provide the performance plots of various representation learning and RL algorithms for the 8M and the 25M benchmarks, which test zero-shot generalization under different sample regimes. Figure~\ref{fig:train_curves_8M} shows the training performance of agents on 8M frames, Figure~\ref{fig:eval_curves_8M} the test performance of agents on 8M frames and all levels, and, finally, Figure~\ref{fig:eval_curves_25M} shows the test performance of agents on 25M frames and all levels.

\paragraph{Ablations on algorithm components}
\label{sec:ablations_algo}
We ran multiple versions of the algorithm to identify the key components which make \shortmethod~ perform well in Procgen. The first modification, \textbf{\shortmethod~consecutive T} consists in running our algorithm but sampling consecutive timesteps, that is $t_{i+1}=t_1+i$ for all $t_1$ and $0\leq i \leq T$. The second modification, \textbf{\shortmethod~no action} removes the action conditioning layer in the log-softmax probability $p_t$ and in the cluster scores $q_t$, to test the importance of action information for cluster membership prediction. The third modification, \textbf{\shortmethod~no cluster} removes the clustering loss, and only restricts to mining and predicting nearby neighbors in the batch. Finally, the last modification, \textbf{\shortmethod~no pred} removes the loss predicting samples from neighboring partitions and only relies on the clustering loss to update its representation.
\begin{table}[ht]
\centering
\caption{\small Average evaluation returns collected after 8M of training frames, $\pm$ one standard deviation.}
\resizebox{\linewidth}{!}{%
\begin{tabular}{l||cccc|c}
\toprule
Env & \shortmethod~consecutive T & \shortmethod~no action & \shortmethod~no cluster & \shortmethod~no pred & \shortmethod{} \\ 
\midrule
bigfish & 3.9$\pm$0.3 & 3.2$\pm$0.3 & 2.5$\pm$0.3 & 3.7$\pm$0.1 & 4.7$\pm$0.2\\ 
bossfight & 8.9$\pm$0.1  & 6.9$\pm$0.9 & 7.8$\pm$0.3 & 6.6$\pm$0.8 & 8.2$\pm$0.1\\ 
caveflyer & 4.6$\pm$0.2 & 4.7$\pm$0.1 & 4.6$\pm$0.1 & 4.6$\pm$0.1 & 4.7$\pm$0.2\\ 
chaser & 7.4$\pm$0.3 & 6.7$\pm$0.2 & 7.0$\pm$0.5 & 6.5$\pm$0.1 & 7.1$\pm$0.2\\ 
climber & 6.2$\pm$0.4 & 5.5$\pm$0.1 & 5.3$\pm$0.4 & 5.7$\pm$0.4 & 5.9$\pm$0.2\\ 
coinrun & 8.8$\pm$0.1 & 8.5$\pm$0.3 & 8.1$\pm$0.2 & 8.4$\pm$0.3 & 8.7$\pm$0.3\\ 
dodgeball & 1.8$\pm$0.1 & 1.7$\pm$0.1 & 1.7$\pm$0.2 & 1.7$\pm$0.1 & 1.8$\pm$0.1\\ 
fruitbot & 13.1$\pm$0.3 & 12.9$\pm$0.4 & 13.0$\pm$0.5 & 12.5$\pm$0.6 & 13.3$\pm$0.3\\ 
heist & 3.0$\pm$0.1 & 3.2$\pm$0.3 & 3.2$\pm$0.1 & 3.0$\pm$0.2 & 3.1$\pm$0.3\\ 
jumper & 6.1$\pm$0.2 & 6.0$\pm$0.1 & 5.9$\pm$0.1 & 5.9$\pm$0.1 & 6.0$\pm$0.1\\ 
leaper & 3.4$\pm$1.1 & 3.2$\pm$0.4 & 2.6$\pm$0.3 & 3.3$\pm$0.2 & 2.8$\pm$0.2\\ 
maze & 5.6$\pm$0.2 & 5.6$\pm$0.1 & 5.7$\pm$0.1 & 5.8$\pm$0.1 & 5.7$\pm$0.1\\ 
miner & 7.0$\pm$0.9 & 5.9$\pm$0.4 & 5.6$\pm$0.1 & 6.0$\pm$0.2 & 6.5$\pm$0.2\\ 
ninja & 5.7$\pm$0.1 & 5.3$\pm$0.2 & 5.5$\pm$0.1 & 5.6$\pm$0.1 &5.8$\pm$0.1 \\ 
plunder & 6.4$\pm$0.2 & 5.6$\pm$0.1 & 5.9$\pm$0.5 & 6.1$\pm$0.2 & 6.6$\pm$0.3\\ 
starpilot & 7.0$\pm$0.2 & 4.9$\pm$0.6 & 4.9$\pm$0.2 & 5.8$\pm$0.4 & 7.7$\pm$0.5\\ \bottomrule
\end{tabular}%
}
\label{tab:procgen_ablation_components}
\end{table}%

Results suggest that (1) using consecutive timesteps as for the dynamics vector embedding yields lower average rewards than non-consecutive timesteps, (2) action conditioning helps the agent to pick up on the local dynamics present in the MDP and (3) both clustering and predictive objectives are essential to the good performance of our algorithm. Results of the last column are computed over 10 seeds, rest over 3 seeds.

\paragraph{Ablation on number of clusters and clustering timesteps}

How should one determine the optimal number of clusters in a complex domain? Can the number of clusters be chosen \emph{a priori} running any training?

Below, we provide some partial answers to these questions. First, the optimal (or true) number of clusters is domain-specific, as it depends on the exact connectivity structure of the MDP at hand. Second, the length of the clusters, i.e. the number of trajectory timesteps passed to Sinkhorn-Knopp can widely impact the nature of learned representations, and hence the downstream performance of the agent.

\begin{figure}[h!]
    \centering
    \includegraphics[width=\linewidth]{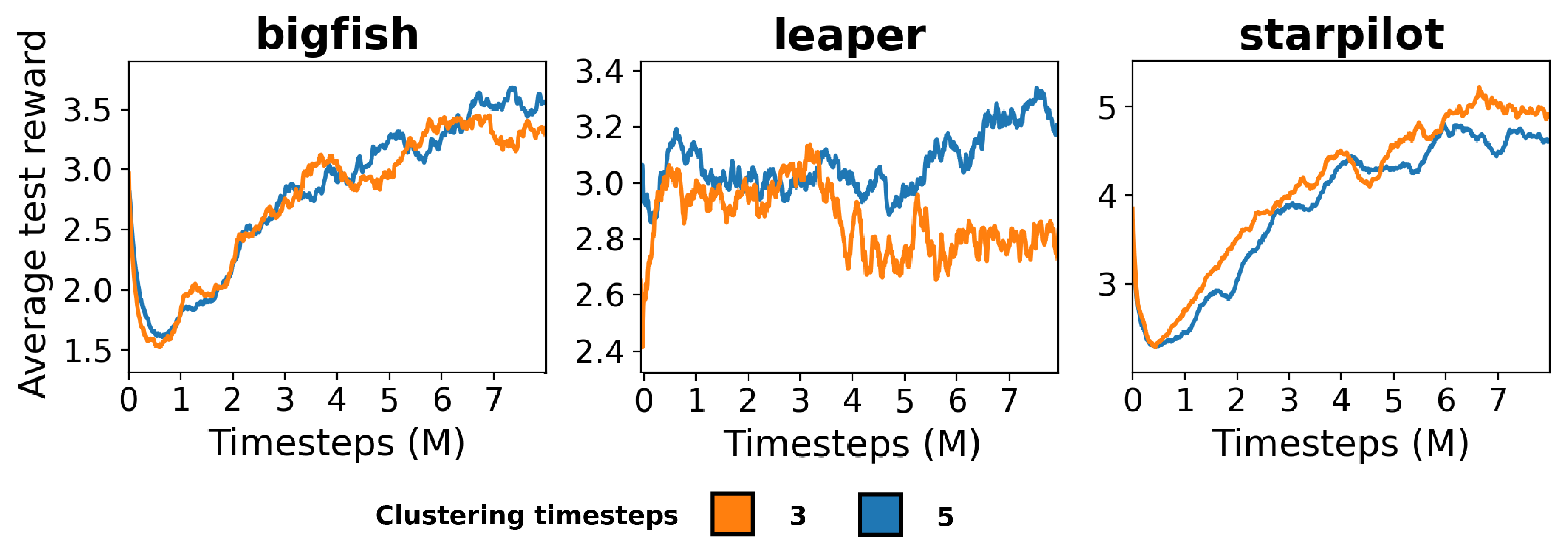}
    \caption{\small Ablation on the clustering timesteps used in the dynamics embedding}
    \label{fig:ablation_cluster_timesteps}
\end{figure}

\begin{figure}[h!]
    \centering
    \includegraphics[width=\linewidth]{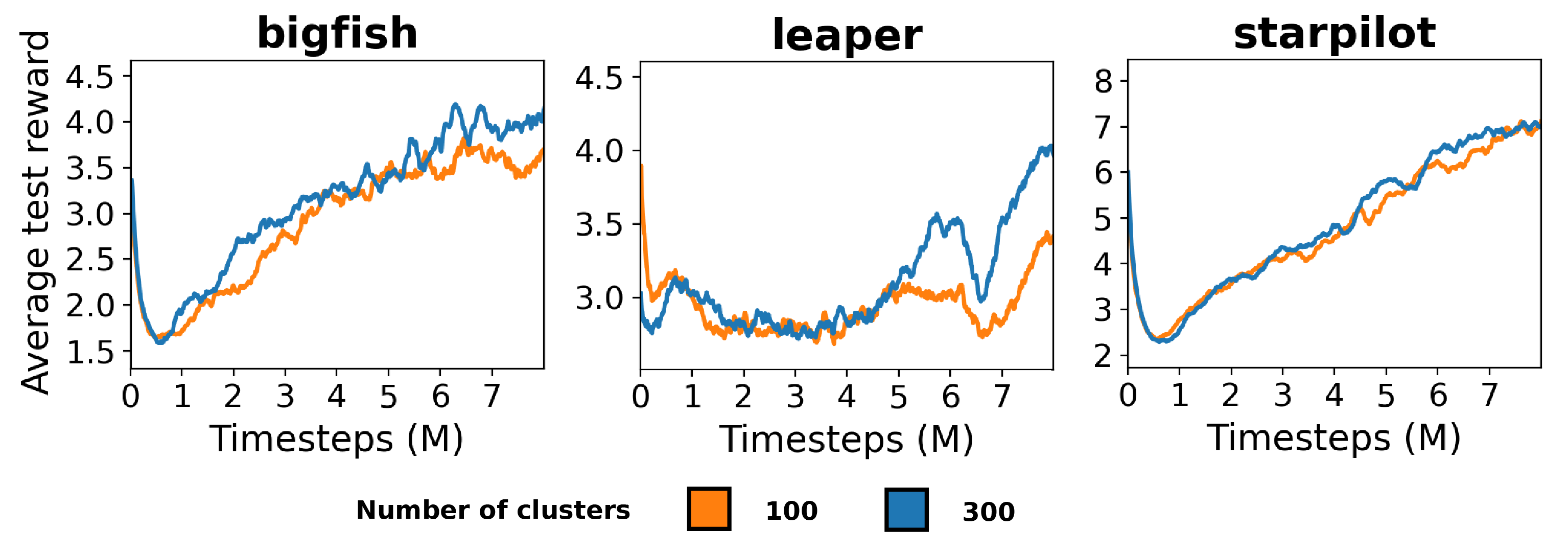}
    \caption{\small Ablation on the number of clusters used in~\shortmethod}
    \label{fig:ablation_n_cluster}
\end{figure}

\paragraph{Temporal connectivity of the clusters}
Are clusters \emph{consistent} in time for a given trajectory? To verify this, we trained \shortmethod~ on 1 million frames of the \textit{bigfish} game. For every $T$ states in a given trajectory, we have computed the hard cluster assignment to the nearest cluster, which yields a sequence of partitions. We then computed the cosine similarity between time-adjacent cluster centroids, the metric reported on the smoothed graph below.

\begin{figure}[h!]
    \centering
    \includegraphics[width=\linewidth]{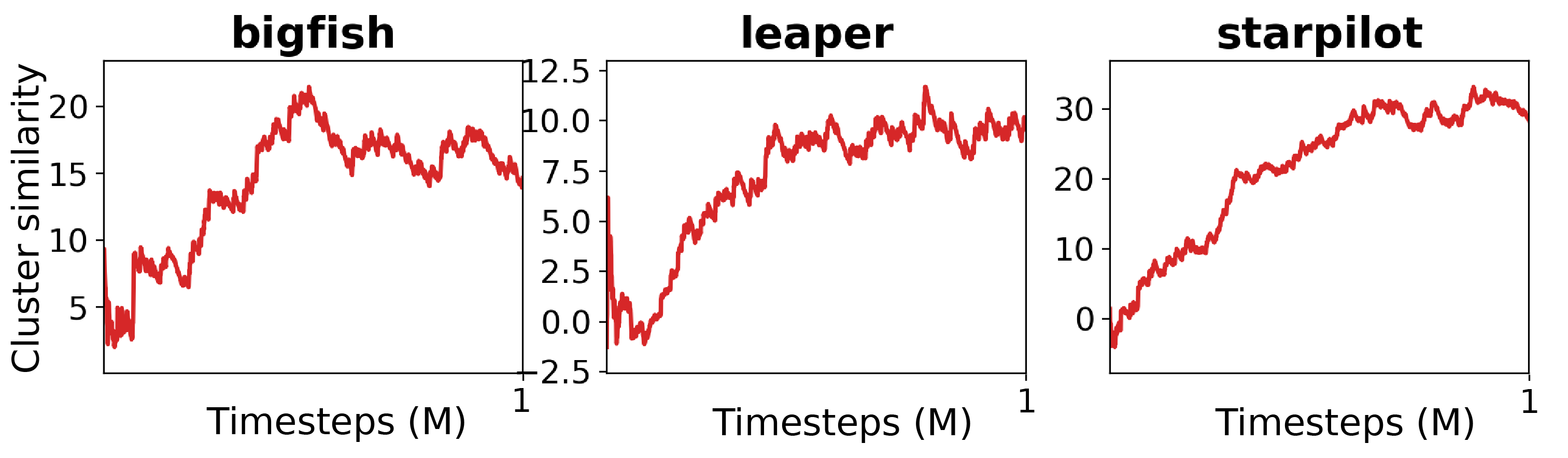}
    \caption{\small Average within-trajectory cluster similarity over 1M consecutive timesteps.}
    \label{fig:ablation_cluster_sim}
\end{figure}

\paragraph{Loss landscape of $\mathcal{L}_{\text{clust}}$ and $\mathcal{L}_{\text{pred}}$}
Works relying on non-colinear signals, e.g. behavioral similarity and rewards, as is the case for DeepMDP~\citep{gelada2019deepmdp}, show that interference can occur between various loss components. For example,~\citep{gelada2019deepmdp} showed how their dynamics and reward losses are inversely proportional to each other early on in the training, taking a considerable amount of frames to converge.
\begin{figure}
    \centering
    \includegraphics[width=0.8\linewidth]{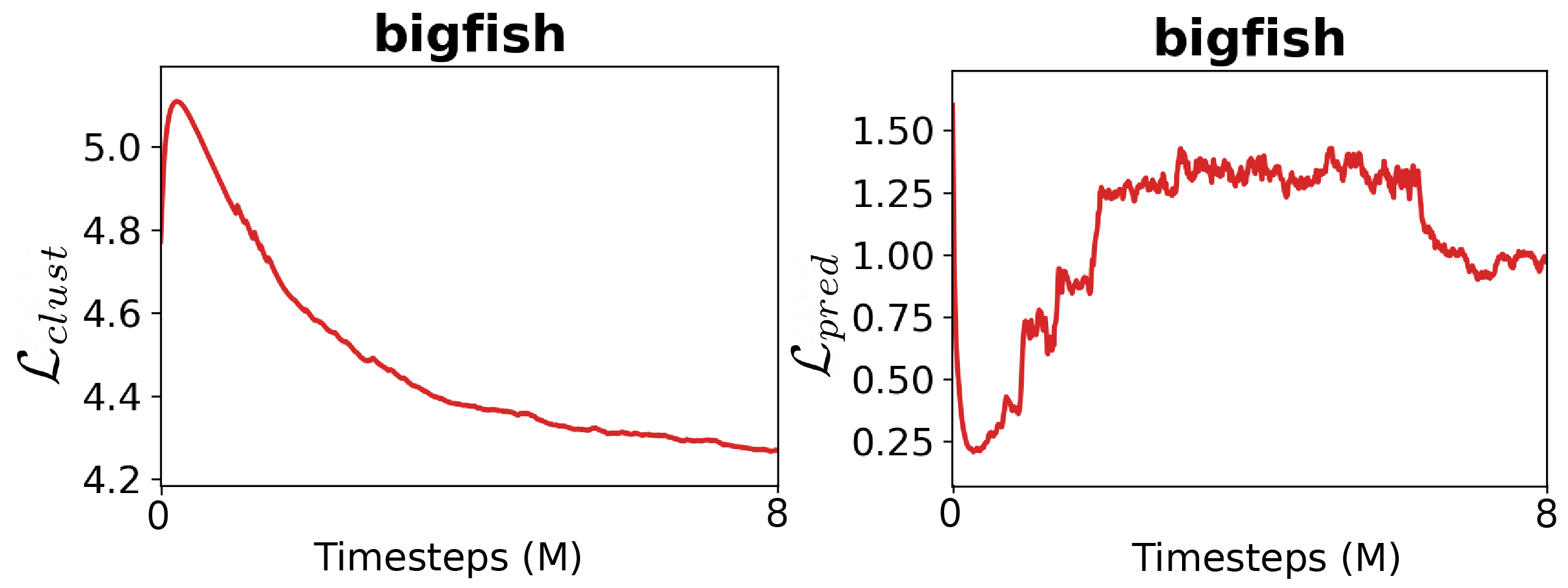}
    \caption{\small Average values of $\mathcal{L}_{clust}$ and $\mathcal{L}_{pred}$ over time. }
    \label{fig:loss_ablation}
\end{figure}

We observe a similar pattern in \Cref{fig:loss_ablation}: the clustering loss first jumps up while the predictive loss is minimized, then the trend reverses, and both losses get minimized near the end of the training.

\paragraph{Case study: splitting other dynamics-aware losses}

Similar to the postulate of ATC~\citep{stooke2020decoupling}, we hypothesize that training the encoder only with the representation loss has the most beneficial effect when the representation loss contains information about dynamics. To validate this, we conducted an additional set of experiments on two well-known self-supervised learning algorithms which leverage predictive information about future timesteps: Deep Reinforcement and InfoMax Learning (DRIML)~\citep{mazoure2020deep} and Self-Predictive Representations (SPR)~\citep{schwarzer2020dataefficient}. We ran (i) the default version of the algorithms with joint RL and representation updates, as well as (ii) RL updates propagated only through the layers above the encoder. 

\begin{table}[h!]
    \centering
    \begin{tabular}{l||ll}
\toprule
       Env & DRIML &    SPR \\
\midrule
   bigfish &  \textbf{+0.17} &   \textbf{+0.15} \\
 bossfight & \textbf{ +0.56} &  \textbf{+10.36} \\
 caveflyer &  \textbf{+0.23} &  -0.02 \\
    chaser & -0.14 &  -0.13 \\
   climber &  \textbf{0.15} &   \textbf{+0.14} \\
   coinrun & -0.37 &   \textbf{+2.05} \\
 dodgeball & -0.39 &   \textbf{+0.12} \\
  fruitbot & -0.53 &   -0.1 \\
     heist &  -0.3 &  -0.03 \\
    jumper &  \textbf{+0.04} &     +0 \\
    leaper & -0.04 &   \textbf{+0.08} \\
      maze &  \textbf{+0.01} &  -0.15 \\
     miner &  \textbf{+0.88} &   \textbf{+0.03} \\
     ninja & -0.13 &  -0.31 \\
   plunder & -0.02 &  -0.18 \\
 starpilot & -0.11 &  -0.15 \\
 \midrule 
 Norm. score & +0.01 & +0.74 \\
\bottomrule
\end{tabular}
    \caption{\small Normalized improvement scores of split updates over joint updates of the encoder, averaged over 3 random seeds.}
    \label{tab:spr_driml}
\end{table}

\begin{figure}[ht]
    \centering
    \includegraphics[width=\linewidth]{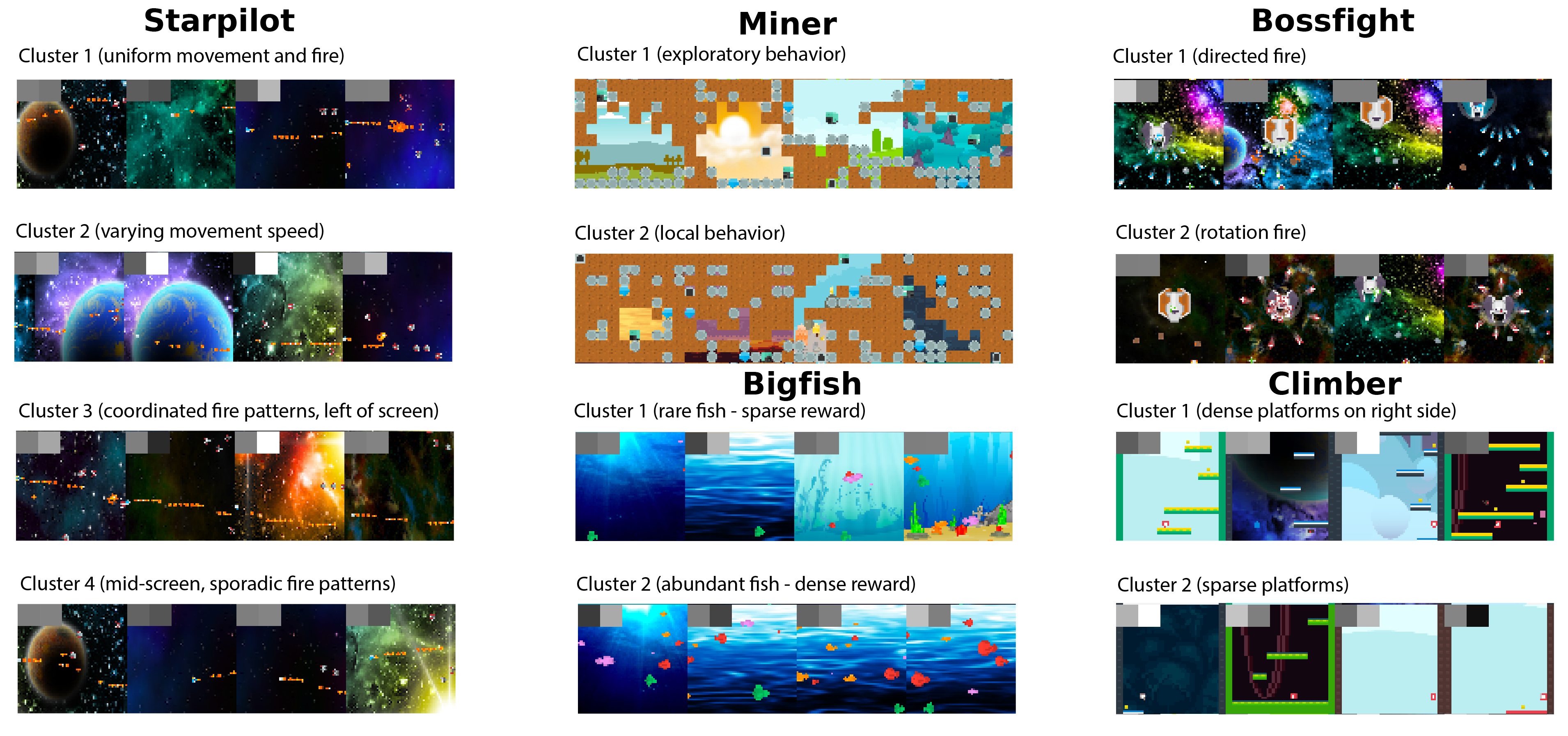}
    \caption{\small Sample states from behavioral clusters found by CTRL after 2M of training frames for 5 representative environments. The two gray squares in top left is added to indicate the agent's velocity.}
    \label{fig:cluster_samples}
\end{figure}

\paragraph{Qualitative assessment of clusters}
Figure~\ref{fig:cluster_samples} shows, for 5 environments, 4 randomly sampled states for 2 behavioral clusters (4 clusters for Starpilot – differences between clusters are easier to visualize in this environment). Note that clustered states go beyond visual similarity, and capture action sequences, agent position, presence of enemies and even topological equivalence of various levels. The choice of environments for this demonstration is dictated by the nature of the action space, e.g. projectiles in Starpilot and path tracing in Miner allow to better visualize agent’s behavior. Note that, for Bigfish, CTRL implicitly picks up the notion of reward density by learning to separate states abundant of fish from those without fish (due to the policy being trained on rewards and thus exhibiting different behavior in those two settings).

\begin{figure}[ht]
    \centering
    \includegraphics[width=0.6\linewidth]{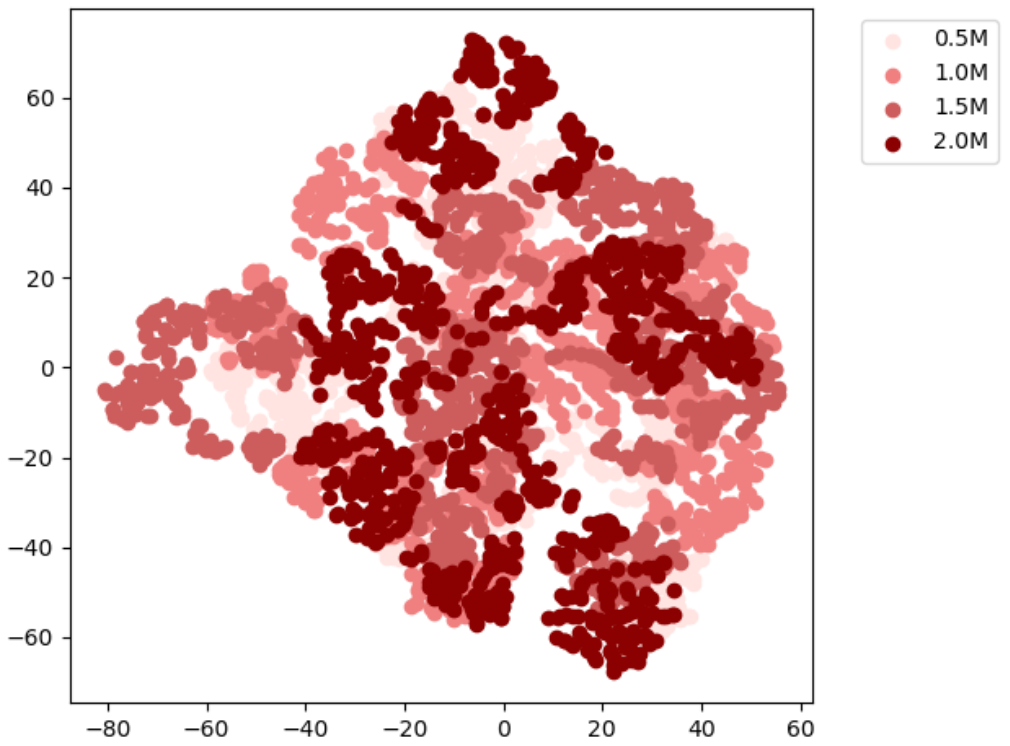}
    \caption{\small t-SNE of 1024 randomly sampled states from data collected by CTRL after 0.5,1,1.5 and 2M frames in Starpilot, with $\beta=0.1, T=4$. As learning progresses, agent behavior clusters become more and more distinct.}
    \label{fig:tsne_clusters}
\end{figure}

Figure~\ref{fig:tsne_clusters} shows the t-SNE of 
 from randomly sampled states along the CTRL training path on Starpilot – embeddings learned by CTRL can be seen to concentrate into distinct clusters and around their respective centroids.

\subsection{Showcase: Slow clustering convergence leads to better generalization\label{sec:slow}}

Trajectory clustering is key to representation learning not only in \shortmethod\ but also in a prior method, Proto-RL. However, while Proto-RL uses it to pretrain a representation which it then keeps frozen during RL, \shortmethod\ applies clustering to evolve the representation as RL progresses. This raises a question: how important is online clustering convergence rate for learning a good representation?
Intuitively, if online clustering converges too quickly and behavioral similarities are ``pinned down'' early in the training process, the resulting representation will not be robust to distribution shifts induced by improved policies. 
Therefore, it seems crucial to learn the behavioral similarities at a rate that allows cluster centroids to adapt to the value improvement path~\citep{dabney2021valueimprovement}. To validate this hypothesis, we conducted an analysis of the correlation between the clustering quality and the test performance as a function of training progress on three representative games.

To measure the clustering quality, we report the silhouette score~\citep{rousseeuw1987silhouettes}, a commonly used unsupervised goodness-of-fit measure which balances inter- and intra-cluster variance. 
\begin{figure}[h!]
    \centering
    \includegraphics[width=0.7\linewidth]{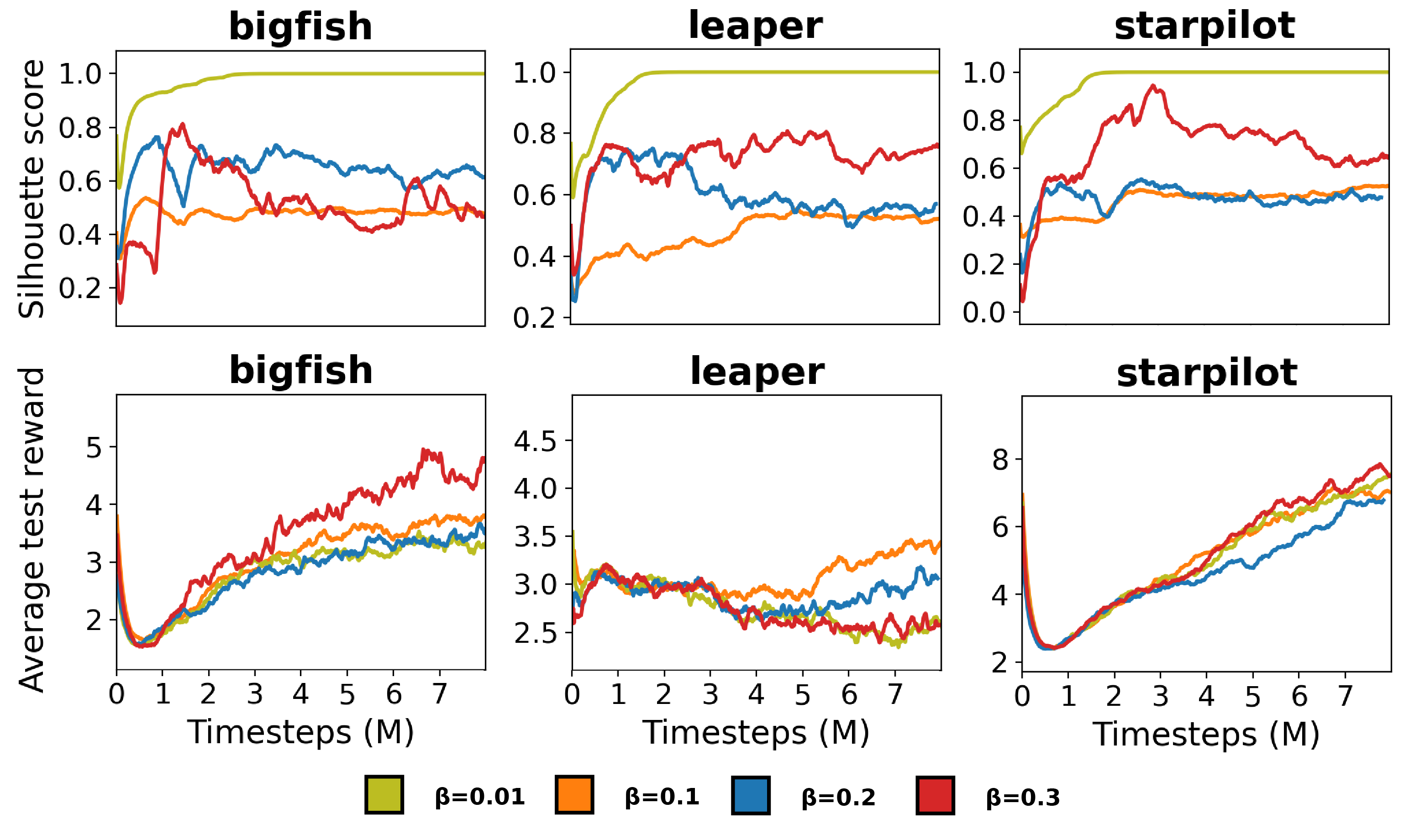}
    \caption{\small Goodness-of-clustering measured by silhouette scores (top) and average test returns (bottom) as a function of training samples.}
    \label{fig:ablation_silhouette}
\end{figure}
Results shown in~\Cref{fig:ablation_silhouette} provide evidence that picking the unsupervised learning procedure which converges the fastest (i.e. uses the lowest temperature $\beta$) does not necessarily lead to the best generalization performance. Based on \Cref{fig:ablation_silhouette}, we conjecture that fast clustering convergence hinders the performance of the RL agent due to clusters being fixed early on and not adapting to the distribution shift induced by the evolving RL policy. 

\subsection{Showcase: Learning behavioral similarities captures local perceptual changes \label{sec:behsim}}

To demonstrate the importance of identifying behavioral similarities, we designed a toy example problem with 5 behavioral clusters, where clustering the behaviors correctly leads to finding a near-optimal policy.

\begin{figure}[h!]
    \centering
    \includegraphics[width=\linewidth]{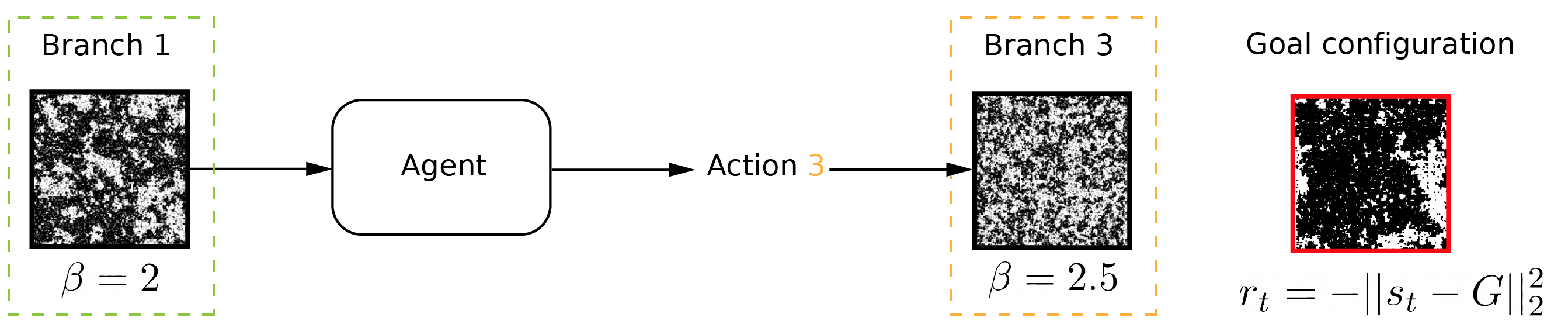}
    \caption{\small The composite Ising matching problem: the agent has to match a given Ising configuration by swapping branches of various transition dynamics}
    \label{fig:ising_schema}
\end{figure}

Our example problem is based on the standard Ising model\footnote{\url{https://en.wikipedia.org/wiki/Ising_model}} -- a $32\times32$ binary lattice, each entry of which evolves at every timestep according to the values of its neighbors, with strengths of neighbor dependencies being regulated by a temperature parameter $1/\beta$. We randomly initialize $5$ Ising models, each parametrized by an inverse temperature parameter on a uniform grid $\beta\in [0.01,0.3]$. The system state is given by the state of all 5 models, and all models evolve in parallel at every step. At each timestep, the agent needs to choose one of the models, and has 5 actions corresponding to these choices. At timestep $t$, the agent is allowed to observe only the state of the model it chose at this step, and gets a reward based only on this model's state. The reward yielded by Ising model $i$ at timestep $t$ is given by $r_t=-||s_{i,t}-G||_2^2$, where $G$ is a goal state. For a given problem instance, we sample $G$ randomly by instantiating a 6th Ising model with an unknown inverse temperature parameter $\beta^*$ and letting $G$ be its final configuration after evolving it for a random number of steps. \Cref{fig:ising_schema} outlines the experimental setting for this study case.

The 5 behavioral clusters in our setting correspond to the 5 Ising models. The optimal strategy to solve this task is to 1) identify the Ising model (i.e., the behavioral cluster) whose temperature parameter is closest to $\beta^*$ and 2) choose that model and collect the corresponding rewards. 

Table~\ref{tab:ising} outlines the results we obtained by deploying \shortmethod\ with different number-of-clusters parameter value. One can see that the largest improvement in silhouette score occurs from $E=4$ to $E=5$ (14.5\%), suggesting that monitoring the largest change in silhouette score can be used to set the true number of clusters in {\shortmethod} which, in turn, corresponds to the highest-return policy discovered by \shortmethod.

\begin{table}[H]
\centering
\begin{tabular}{l|llllll}
\toprule
 & $E=2$ &  $E=4$  & $E=5$ & $E=6$ & \ldots & $E=50$ \\ \midrule
 Returns & -0.78 & -0.96 & -0.02 & -0.07 & \ldots & -0.54\\
 Silhouette & 0.875 & 0.796 & 0.651 & 0.554 & \ldots & 0.039\\
 Silhouette change & - & 0.079 & \textbf{0.145} & 0.097 & \ldots & 0.515\\
\bottomrule
\end{tabular}%
\caption{\small Returns and silhouette scores obtained by \shortmethod~in the composite Ising matching domain.}
\label{tab:ising}
\end{table}

\subsection{Additional theoretical findings}

\paragraph{Do uncorrelated local changes to state embeddings affect the clustering? }

\begin{theorem}
Let $M$ be an MDP and let $\vec{v}\in\mathcal{V}$ be a dynamics embeddings in $M$. Define
\begin{equation}
   \begin{cases}
   \vec{\delta}_{i}=\vec{\delta}_i^1 & 1 \leq i \leq |\mathcal{V}|\\
   \vec{\delta}_{i}=\vec{\delta}_i^2 & h < i \leq T|\mathcal{V}|\\
   \end{cases} 
\end{equation}
and pick $\vec{\delta}^1$ s.t. it lies on the positive half-plane spanned by $\mat{E}^\top_{j}-\mat{E}^\top_{j'}$ for some $1\leq j' \leq E$. Then, $\vec{v}'=\vec{v}+\vec{\delta}$ and $\vec{v}$ belong to the same partition $j$.
\label{thm:directional_perturbation}
\end{theorem}

It becomes apparent from the above statement that perturbations to a single state or groups of state embeddings \emph{do not} modify the partition membership as long as their direction aligns with that of the cluster embeddings.

\subsection{Proofs}\label{app:proofs}
Throughout this section, we assume that the policy $\pi$ is fixed, and that {\shortmethod} optimizes $\mathcal{L}_{clust}$ only.

\begin{proof}[Theorem~\ref{thm:directional_perturbation}]
For two dynamics embeddings to be assigned to the same cluster $j$, the following should hold
\begin{equation}
\begin{split}
    \sum_{i=1}^{|\mathcal{V}|}\vec{v}_{i}\mat{E}^\top_{ji}&>\sum_{i=1}^{|\mathcal{V}|}\vec{v}_{i}\mat{E}^\top_{j'i},\\
    \sum_{i=1}^{|\mathcal{V}|}(\vec{v}_{i}+\vec{\delta}_i)\mat{E}^\top_{ji}&>\sum_{i=1}^{|\mathcal{V}|}(\vec{v}_{i}+\vec{\delta}_i)\mat{E}^\top_{j'i}
\end{split}
\end{equation}
for any $1\leq j' \leq E$ s.t. $j'\neq j$.

\begin{equation}
    \begin{split}
        \sum_{i=1}^{|\mathcal{V}|}\vec{v}_{i}\mat{E}^\top_{ji}+\sum_{i=1}^{|\mathcal{V}|}\vec{\delta}_i\mat{E}^\top_{ji}&>\sum_{i=1}^{|\mathcal{V}|}\vec{v}_{i}\mat{E}^\top_{j'i}+\sum_{i=1}^{|\mathcal{V}|}\vec{\delta}_i\mat{E}^\top_{j'i}\\
        \sum_{i=1}^{|\mathcal{V}|}\vec{v}_{i}(\mat{E}^\top_{ji}-\mat{E}^\top_{j'i})+\sum_{i=1}^{|\mathcal{V}|}\vec{\delta}_i(\mat{E}^\top_{ji}-\mat{E}^\top_{j'i})&>0
    \end{split}
\end{equation}

Taking the difference between both equations yields the necessary condition for two dynamics to belong to the same cluster
\begin{equation}
   \sup_{1\leq j' \leq E} (\mat{E}^\top_{ji}-\mat{E}^\top_{j'i})\vec{\delta}_i\geq 0,  \; 1 \leq i \leq |\mathcal{V}|.
   \label{eq:cluster_assignment_condition}
\end{equation}

\begin{corollary}
Let $\vec{v},\vec{v}'$ be two dynamics embeddings, and define $\vec{\delta}=\vec{v}'-\vec{v}$. If $\vec{v}$ belongs to cluster $j$ and $j=\argmax_{1\leq j' \leq E}\mat{E}^\top_{j'}\vec{\delta}$, then $\vec{v}'$ also belongs to cluster $j$.
\end{corollary}

Perturbations are of the form $\sum_{i=1}^{|\mathcal{V}|}(\mat{E}^\top_{ij}-\mat{E}^\top_{ij'})\vec{\delta}_i$. If $\vec{\delta}=0$, then the cluster assignment doesn't change. Let $\vec{v}$ be of size $k h=|\mathcal{V}|$. Define, without loss of generality
\begin{equation}
   \begin{cases}
   \vec{\delta}_{i}=\vec{\delta}_i^1 & 1 \leq i \leq h\\
   \vec{\delta}_{i}=\vec{\delta}_i^2 & h < i \leq kh\\
   \end{cases} 
\end{equation}
and pick $\vec{\delta}^1$ s.t. it lies on the positive half-plane spanned by $\mat{E}^\top_{ij}-\mat{E}^\top_{ij'}$.

Then,
\begin{equation}
    \sum_{i=1}^{|\mathcal{V}|}(\mat{E}^\top_{ij}-\mat{E}^\top_{ij'})\vec{\delta}_i=\sum_{i=1}^{h}(\mat{E}^\top_{ij}-\mat{E}^\top_{ij'})\vec{\delta}_i^1+\sum_{i=h}^{kh}(\mat{E}^\top_{ij}-\mat{E}^\top_{ij'})\vec{\delta}_i^2\geq \sum_{i=h}^{kh}(\mat{E}^\top_{ij}-\mat{E}^\top_{ij'})\vec{\delta}_i^2\geq 0
\end{equation}

which concludes the proof.
\end{proof}

\begin{proof}[Theorem~\ref{thm:bisimulation_partition}]
Since the $\mathcal{W}_1$ metric is defined between distribution functions, we use $\vec{v}=\mathbb{P}[\vec{v}]$ throughout the proof to denote the probability distribution over elements of the dynamics vector $\vec{v}$. In practice, this amounts to re-normalizing the representation.

For two dynamics to be assigned to the same cluster $j$, the following has to hold:
\begin{equation}
\begin{split}
    \sum_{i=1}^{|\mathcal{V}|}\vec{v}_{i}\mat{E}^\top_{ji}&>\sum_{i=1}^{|\mathcal{V}|}\vec{v}_{i}\mat{E}^\top_{j'i},\\
    \sum_{i=1}^{|\mathcal{V}|}\vec{v}_{i}'\mat{E}^\top_{ji}&>\sum_{i=1}^{|\mathcal{V}|}\vec{v}_{i}'\mat{E}^\top_{j'i}
\end{split}
\end{equation}
for any $1\leq j' \leq E$ s.t. $j'\neq j$. Then, adding both inequalities yields, for all $1\leq j \leq E$

\begin{equation}
    \begin{split}
        \sum_{i=1}^{|\mathcal{V}|}\vec{v}_{i}\mat{E}^\top_{ji}+\sum_{i=1}^{|\mathcal{V}|}\vec{v}_{i}'\mat{E}^\top_{ji}&\geq\sum_{i=1}^{|\mathcal{V}|}\vec{v}_{i}\mat{E}^\top_{j'i}+\sum_{i=1}^{|\mathcal{V}|}\vec{v}_{i}'\mat{E}^\top_{j'i}\\
         \sum_{i=1}^{|\mathcal{V}|}\vec{v}_{i}\mat{E}^\top_{ji}+\sum_{i=1}^{|\mathcal{V}|}\vec{v}_{i}\mat{E}^\top_{j'i}&\geq\sum_{i=1}^{|\mathcal{V}|}\vec{v}_{i}'\mat{E}^\top_{ji}+\sum_{i=1}^{|\mathcal{V}|}\vec{v}_{i}'\mat{E}^\top_{j'i}\\
         \sum_{i=1}^{|\mathcal{V}|}\vec{v}_{i}(\mat{E}^\top_{ji}-\mat{E}^\top_{j'i})&\geq\sum_{i=1}^{|\mathcal{V}|}\vec{v}_{i}'(\mat{E}^\top_{ji}-\mat{E}^\top_{j'i})\\
         \sum_{i=1}^{|\mathcal{V}|}\vec{v}_{i}(\mat{E}^\top_{ji}-\mat{E}^\top_{j'i})&\geq\sum_{i=1}^{|\mathcal{V}|}\vec{v}_{i}'(\mat{E}^\top_{ji}-\mat{E}^\top_{j'i})\\
    \end{split}
\end{equation}

and the constraint of two vectors belonging to the same cluster $j$ becomes
\begin{equation}
    \begin{split}
         \sum_{i=1}^{|\mathcal{V}|}(\vec{v}_{i}-\vec{v}_{i}')(\mat{E}^\top_{ji}-\mat{E}^\top_{j'i})&\geq 0\\
         \min_{1\leq j' \leq E} \sum_{i=1}^{|\mathcal{V}|}(\vec{v}_{i}-\vec{v}_{i}')(\mat{E}^\top_{ji}-\mat{E}^\top_{j'i})&\geq 0\\
         \min_{1\leq j' \leq E} (\vec{v}-\vec{v}')(\mat{E}_{j}-\mat{E}_{j'})^\top &\geq 0\\
    \end{split}
\end{equation}

 Now, denote $\mat{E}(j):=\mat{E}_j$. Our constraint satisfaction problem can be written as
\begin{equation}
    \min_{1\leq j' \leq E} (\vec{v}-\vec{v}')(\mat{E}(j)-\mat{E}(j'))^\top \geq 0
    \label{eq:cluster_mat_form}
\end{equation}

By comparing Eq.~\ref{eq:bisimulation_mat_form} with Eq.~\ref{eq:cluster_mat_form}, we observe that in our case, $\mu$ is restricted to the set of vectors in $\mathbb{R}^{|\mathcal{V}|}$. Therefore, we pick
$\mu\in \Gamma(\mat{E})$, where $\Gamma(\mat{E})=\{\vec{\omega}\in \mathcal{V}:\vec{\omega}= \mat{E}(j,i)-\mat{E}(j',i), 0\leq  \vec{\omega}_i\leq 1, \cos(\vec{v}-\vec{v}',\vec{\omega})\in [0,\pi] | 1 \leq i \leq |\mathcal{V}|, 1 \leq j' \leq E\}$. The set $\Gamma(\mat{E})$ is non-empty if $\max_{l,l'}||\mat{E}_{l}-\mat{E}_{l'}||_\infty \leq 1$, which holds due to $\ell_p$ norm ordering \emph{and} since $\mat{E}$ is normalized in the $\tilde{\mat{Q}}$ scores expression. Adopting this notation simplifies the previous expression to

\begin{equation}
    \min_{\mu\in \Gamma(\mat{E})} (\vec{v}-\vec{v}')\mu^\top 
    \label{eq:cluster_mu}
\end{equation}

Once again, recall that $\mat{E}$ is normalized. Therefore, we have
\begin{equation}
    \begin{split}
        \bigg(\frac{\vec{e}_{ij}}{||\vec{e}_j||_2}-\frac{\vec{e}_{i'j}}{||\vec{e}_j||_2}\bigg)+ \bigg(\frac{\vec{e}_{ij}}{||\vec{e}_j||_2}-\frac{\vec{e}_{i'j'}}{||\vec{e}_j||_2}\bigg) & \leq d(i,i')\\
    \end{split}
\end{equation}

which equivalently can be re-stated as (for $e_i\geq e_j$ WLOG):
\begin{equation}
    \begin{split}
        \vec{e}_{ij}-\vec{e}_{i'j} & \leq \frac{||\vec{e}_j||_2}{2} d(i,i')\\
         \vec{e}_{ij}-\vec{e}_{i'j} & \leq \frac{||\vec{e}_j||_2^2}{2}\big(\mat{E}_{ij}-\mat{E}_{ij'}\big)\\
         \vec{e}_{ij}-\vec{e}_{i'j} & \leq \frac{||\vec{e}_j||_2}{2}\big(\vec{e}_{ij}-\vec{e}_{i'j}\big),\\
    \end{split}
\end{equation}

where we take, as an example, $d(i,i')=||\vec{e}_j||_2\big(\mat{E}_{ij}-\mat{E}_{ij'}\big)$.

The final expression for the sufficient condition for two dynamics embeddings to belong to the same partition is
\begin{equation}
    \begin{split}
        \min_{\mu\in \Gamma(\mat{E})} (\vec{v}-\vec{v}')\mu^\top \\
        \text{s.t.}\; \mu(i)-\mu(i')\leq d(i,i')
    \end{split}
\end{equation}
for $d(i,i')=||\vec{e}_j||_2\big(\mat{E}_{ij}-\mat{E}_{ij'}\big)$, which is similar to the Wasserstein-1 distance under $d$, i.e. $\mathcal{W}_1^d(\vec{v},\vec{v}')$.

We constructed an operator similar to $\mathcal{F}(d)$ in \cite{ferns2004metrics}. $d$ can be computed by recursively applying $\mathcal{F}(d)$ at each $\vec{v},\vec{v}'\in\mathcal{V}$ pointwise, which is similar to what is done in {\shortmethod}. This concludes our proof and shows how our clustering procedure can be viewed as finding reward-free bisimulations.
\end{proof}

However, note that the exact interpretation of \emph{reward-free bisimulation relation} depends on how $\vec{v}$ is defined. Taking $\vec{v}$ to be two consecutive timesteps of state-action pairs yields the closest possible to the original definition of bisimulation, while sampling temporal keypoints far across the trajectory will induce a different set of properties.

\end{document}